%% file: main.tex
\documentclass[acmsmall,nonacm]{acmart}
\AtBeginDocument{%
  \providecommand\BibTeX{{%
    \normalfont B\kern-0.5em{\scshape i\kern-0.25em b}\kern-0.8em\TeX}}}

\setcopyright{none}
\renewcommand\footnotetextcopyrightpermission[1]{} %

\usepackage{adjustbox}
\usepackage{footnote}
\usepackage{longtable}
\makesavenoteenv{table}

\usepackage{subfigure}

\usepackage{smartdiagram}
\usepackage{tikz}
\usetikzlibrary{shapes,arrows}
\usepackage[edges]{forest}
\usepackage{url}

\usepackage{ulem}
\usepackage{xcolor}

\newcommand{\appref}[1]{Appendix~\ref{#1}}

\begin{document}

\title{Data-Centric Foundation Models in Computational Healthcare: A Survey}

\author{Yunkun Zhang}
\authornote{\ Equal Contribution \quad $^{\dagger}$ Corresponding Author: zhang.shaoting@sjtu.edu.cn, dequanwang@sjtu.edu.cn}
\affiliation{%
  \institution{Shanghai Jiao Tong University}
  \city{Shanghai}
  \country{China}
}

\author{Jin Gao}
\authornotemark[1]
\affiliation{
  \institution{Shanghai Jiao Tong University}
  \city{Shanghai}
  \country{China}
}

\author{Zheling Tan}
\affiliation{
  \institution{Shanghai Jiao Tong University}
  \city{Shanghai}
  \country{China}
}

\author{Lingfeng Zhou}
\affiliation{
  \institution{Shanghai Jiao Tong University}
  \city{Shanghai}
  \country{China}
}

\author{Kexin Ding}
\affiliation{
  \institution{Rutgers University}
  \state{New Jersey}
  \country{USA}
}

\author{Mu Zhou}
\affiliation{
  \institution{Rutgers University}
  \state{New Jersey}
  \country{USA}
}

\author{Shaoting Zhang}
\email{zhang.shaoting@sjtu.edu.cn}
\authornotemark[2]
\affiliation{
  \institution{Shanghai Jiao Tong University}
  \city{Shanghai}
  \country{China}
}

\author{Dequan Wang}
\authornotemark[2]
\email{dequanwang@sjtu.edu.cn}
\affiliation{
  \institution{Shanghai Jiao Tong University, China, and Shanghai Innovation Institute}
  \city{Shanghai}
  \country{China}
}

\renewcommand{\shortauthors}{Y. Zhang et al.}

\begin{abstract}

The advent of foundation models (FMs) as an emerging suite of AI techniques has struck a wave of opportunities in computational healthcare. The interactive nature of these models, guided by pre-training data and human instructions, has ignited a data-centric AI paradigm that emphasizes better data characterization, quality, and scale. In healthcare AI, obtaining and processing high-quality clinical data records has been a longstanding challenge, encompassing data quantity, annotation, patient privacy, and ethics. In this survey, we investigate a wide range of data-centric approaches in the FM era (from model pre-training to inference) towards improving the healthcare workflow. We discuss key perspectives in AI security, assessment, and alignment with human values. Finally, we offer a promising outlook on FM-based analytics to enhance patient outcomes and clinical workflows in the evolving landscape of healthcare and medicine. We provide an up-to-date list of healthcare-related foundation models and datasets at \url{https://github.com/Yunkun-Zhang/Data-Centric-FM-Healthcare}.

\end{abstract}



\maketitle

\input{content_files/1-introduction}
\input{content_files/2-FM}
\input{content_files/3-HealthcareFM}
\input{content_files/4-modality}
\input{content_files/5-quantity}
\input{content_files/6-annotation}

\input{content_files/7-privacy}
\input{content_files/8-evaluation}
\input{content_files/9-outlook}
\input{content_files/10-conclusion}

\begin{acks}
This work was supported by the Shanghai Agricultural Science and Technology Innovation Program (No. K2025001).
We would like to express our sincere gratitude to Xiaofan Zhang and Shengyi Hua from Shanghai Jiao Tong University for their valuable comments and suggestions on this paper.
\end{acks}

\bibliographystyle{ACM-Reference-Format}
\bibliography{bib_files/1-intro,bib_files/2-FM,bib_files/3-MedFM,bib_files/5-challenge-quality,bib_files/6-challenge-quantity,bib_files/4-challenge-modality,bib_files/7-challenge-annotation,bib_files/8-challenge-privacy,bib_files/9-challenge-ethics,bib_files/10-evaluation,bib_files/11-limitation.bib,bib_files/top70.bib,bib_files/full-literature-extra.bib}

\appendix
\newpage
\input{content_files/appendix.tex}

\end{document}

%% file: content_files/1-introduction.tex
\section{Introduction}

The rise of \emph{foundation models} (FMs) has ushered in a wave of breakthroughs for visual recognition~\cite{radford2021clip,kirillov2023sam,ramesh2021dalle}, language understanding~\cite{devlin2018bert,brown2020gpt3}, instruction-following conversational systems~\cite{chatgpt,openai2023gpt4}, and knowledge discovery~\cite{bommasani2021foundation,pan2023unifying}.
In computational healthcare~\cite{esteva2019guide,aggarwal2021diagnostic}, FMs can handle a variety of clinical data with their appealing capabilities in logical reasoning and semantic understanding. Examples span fields in medical conversation~\cite{MedAlpaca,yunxiang2023chatdoctor}, patient health profiling~\cite{christodoulou2019systematic}, and treatment planning~\cite{nori2023gpt4-on-medical}.
Moreover, given the strength in large-scale data processing, FMs offer a shifting paradigm to assess real-world clinical data in the healthcare workflow rapidly and effectively~\cite{thirunavukarasu2023large,qiu2023survey-ai-health}.
We use \emph{large multimodal models} (LMMs) to denote general-purpose foundation models that natively operate over multiple modalities, and \emph{generalist medical AI} (GMAI) to denote the clinical instantiation aiming to generalize across tasks, modalities, and workflows~\cite{moor2023gmai}.
Beyond general-purpose models, specialty medical vision FMs trained on large real-world datasets have started to demonstrate clinical-grade performance in pathology~\cite{vorontsove2024clinicalgrade_r08} and dermatology~\cite{amultimodalvision2025_803e276d}, with prospective evaluation also emerging in ophthalmology~\cite{aneyecarefoundation2025_e5f2b10e}.
Meanwhile, foundation models are rapidly emerging for imaging specialties beyond the above exemplars, including ultrasound and volumetric radiology~\cite{echofmfoundationmodel2025_45c40046,merlinavision2024_2f35dbf0,gu2025anatomy}.

FM research places a sharp focus on the \emph{data-centric} perspective~\cite{zha2023data-centric}.
First, FMs demonstrate the power of \emph{scale}, where the enlarged model and data size permit FMs to capture vast amounts of information, thus increasing the pressing need of training data quantity~\cite{vaswani2017attention}.
Second, FMs encourage \emph{homogenization}~\cite{bommasani2021foundation} as evidenced by their extensive adaptability to downstream tasks. High-quality data for FM training thus becomes critical since it can impact the performance of both pre-trained FM and downstream models.
Therefore, addressing key data challenges is progressively recognized as a research priority.
In the healthcare system, collecting high-quality records could enable a comprehensive understanding of patient characteristics (imaging, genomics, and lab testing data)~\cite{johnson2023data,singh2023systematic}. As illustrated, data-centric strategies promise to reshape clinical workflow~\cite{rao2023assessing,juluru2021integrating}, enable precise diagnosis~\cite{hunter2022role}, and uncover insights into treatment~\cite{chen2023artificial}.

\input{content_files/figures/overview_figure.tex}

Medical data challenges have posed persistent obstacles over decades, including multi-modality data fusion (Section~\ref{section:modality}), limited data volume (Section~\ref{section:quantity}), annotation burden (Section~\ref{section:annotation}), and the critical concern of patient privacy protection (Section~\ref{section:privacy})~\cite{huang2020fusion,hathaliya2020security-privacy,rajpurkar2022ai-health}.
To respond, the FM era opens up perspectives to advance data-focused AI analytics.
Multi-modal FMs, as a concrete example, can offer scalable data fusion strategies for various data formats~\cite{li2023multimodal,ding2023large}.
Meanwhile, the appealing trait of FMs to generate high-quality synthetic data can help mitigate data scarcity while supporting privacy-preserving data sharing~\cite{tang2023does,chambon2022adapting,lu2022comparative,chang2023mining}.
To build responsible solutions for healthcare AI, the evolving perspective on \emph{AI-human alignment}~\cite{gabriel2020artificial,ngo2022alignment} has become increasingly important.
We discuss the necessity of the real-world applications of FMs aligned with human ethics, equity, and societal norms to reduce potential risks in performance assessment, ethical compliance, and patient safety~\cite{liu2023trustworthy,hathaliya2020security-privacy}.
In the FM era, enabling AI-human alignment further underscores the significance of data focus, motivating us to prioritize the data-centric challenges in the landscape of computational healthcare.

\input{content_files/figures/issues.tex}

In this survey, we offer a scoping perspective on developing, analyzing, and evaluating FM-focused approaches for healthcare.
Differentiated from most previous literature reviews, which mainly discuss FM architectures and applications~\cite{thirunavukarasu2023large,qiu2023survey-ai-health,zhang2023challenges}, our survey emphasizes the interplay between patients, healthcare data, and foundation models from a data-centric viewpoint as seen in Fig.~\ref{fig:overview}.
We collect and discuss essential concepts, models, datasets, and tools for analyzing FMs (Fig.~\ref{fig:issues}). Finally, we highlight emerging risks of applying FMs in healthcare and medicine regarding to privacy protection and ethical use. We offer promising directions for FM-based analytics to enhance the predictive performance of patient outcomes and streamline the clinical data workflows, ultimately leading to building better AI-human-aligned, data-focused tools, approaches, and systems in healthcare and medicine.

%% file: content_files/figures/overview_figure.tex
\begin{figure}[t]
    \centering
    \includegraphics[width=\linewidth]{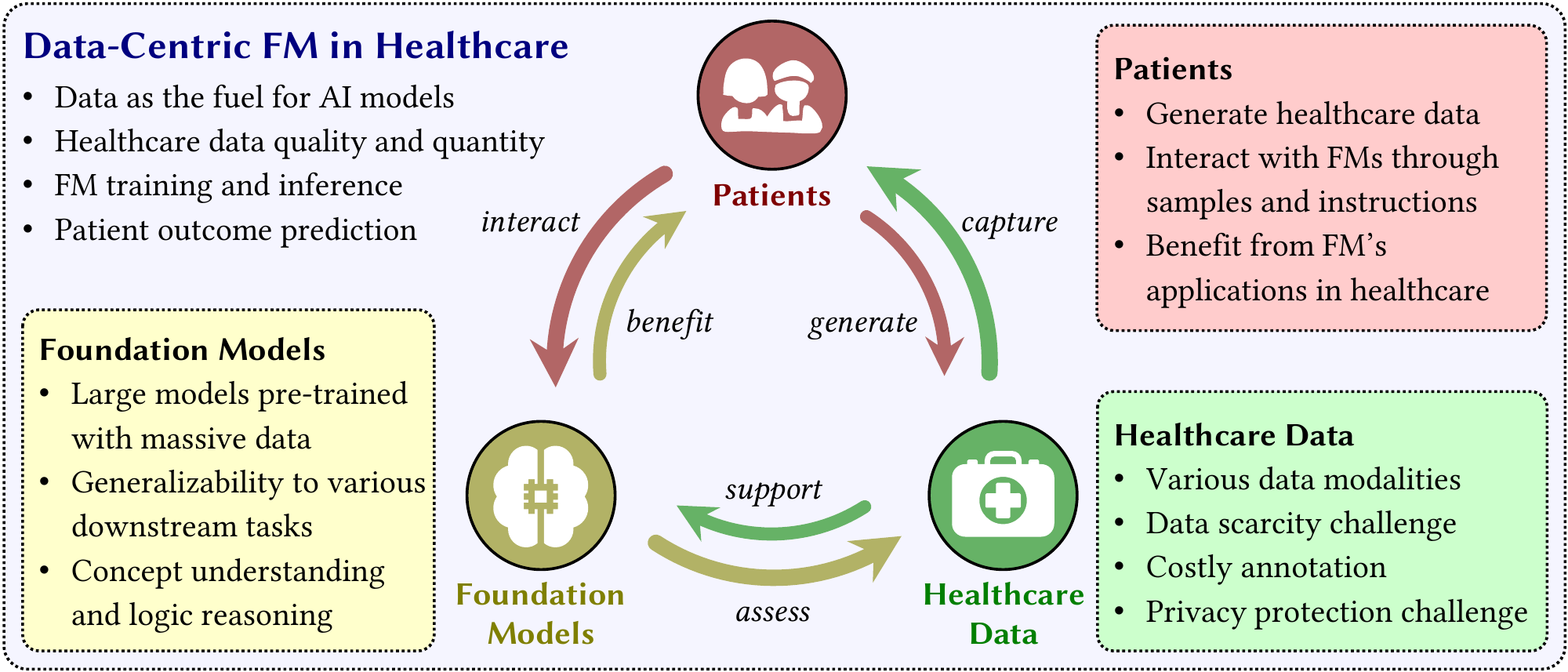}
    \caption{Data-centric foundation models in computational healthcare.}
    \label{fig:overview}
\end{figure}

%% file: content_files/figures/issues.tex
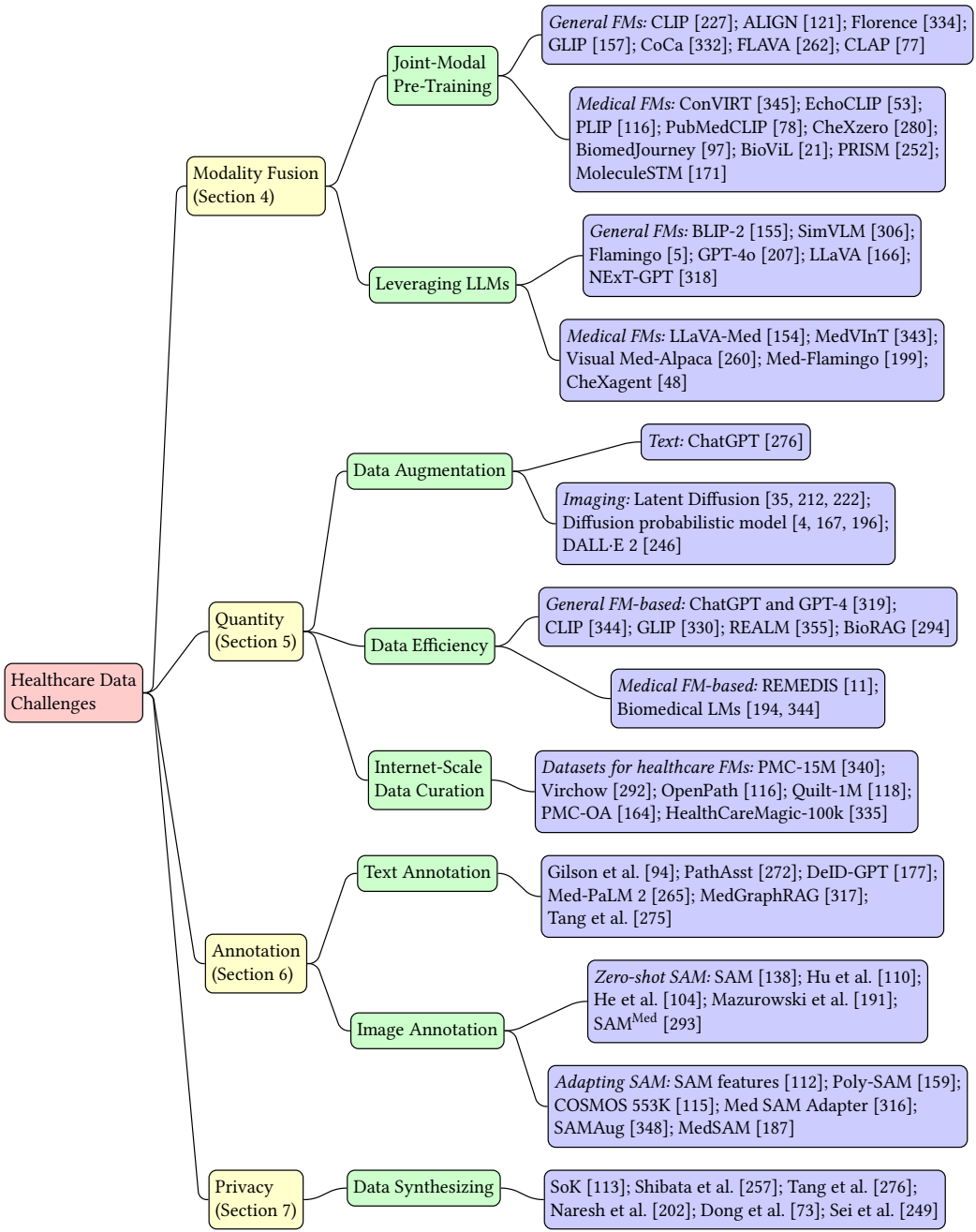
\begin{figure}[t]
\adjustbox{max width=\linewidth}{
\begin{forest}
for tree={%
  draw, semithick, rounded corners, %
  where level = 0{top color = red!20, bottom color = red!20}{},
  where level = 1{top color = yellow!20, bottom color = yellow!20}{},
  where level = 2{top color = green!20, bottom color = green!20}{},
  where level = 3{top color = blue!20, bottom color = blue!20}{},
        align = left,
         edge = {draw, semithick, rounded corners},
         edge path'={
        (!u.parent anchor) -- ($(!u.parent anchor)+(5pt,0pt)$) -- ($(.child anchor)-(5pt,0pt)$) -- (.child anchor)
      },
parent anchor = east, 
 child anchor = west,
        grow' = east,
        s sep = 4mm,    %
        l sep = 8mm,    %
}
[Healthcare Data \\ Challenges
    [Modality Fusion \\ (Section~\ref{section:modality})
        [Joint-Modal \\ Pre-Training
            [\emph{General FMs:} CLIP~\cite{radford2021clip}; ALIGN~\cite{jia2021scaling}; Florence~\cite{yuan2021florence}; \\ GLIP~\cite{li2022grounded}; CoCa~\cite{yu2022coca}; FLAVA~\cite{singh2022flava}; CLAP~\cite{elizalde2023clap}]
            [\emph{Medical FMs:} ConVIRT~\cite{zhang2022contrastive}; EchoCLIP~\cite{christensen2024echoclip}; \\ PLIP~\cite{huang2023plip}; PubMedCLIP~\cite{eslami2021pubmedclip}; CheXzero~\cite{tiu2022expert}; \\ BiomedJourney~\cite{gu2023biomedjourney}; BioViL~\cite{boecking2022making}; PRISM~\cite{shaikovski2024prism}; \\ MoleculeSTM~\cite{liu2023moleculestm}]
        ]
        [Leveraging LLMs
            [\emph{General FMs:} BLIP-2~\cite{li2023blip2}; SimVLM~\cite{wang2021simvlm}; \\ Flamingo~\cite{alayrac2022flamingo}; GPT-4o~\cite{gpt4o}; LLaVA~\cite{liu2023vinst}; \\ NExT-GPT~\cite{wu2023next-gpt}]
            [\emph{Medical FMs:} LLaVA-Med~\cite{li2023llava-med}; MedVInT~\cite{zhang2023pmc}; \\ Visual Med-Alpaca~\cite{MedAlpaca}; Med-Flamingo~\cite{moor2023med-flamingo}; \\ CheXagent~\cite{chen2024chexagent}; PneumoLLM~\cite{song2024pneumollm}]
        ]
    ]
    [Quantity \\ (Section~\ref{section:quantity})
        [Data Augmentation
            [\emph{Text:} ChatGPT~\cite{tang2023does,yuan2024large}]
            [\emph{Imaging:} Latent Diffusion~\cite{pinaya2022brain,chambon2022adapting}; \\ Diffusion probabilistic model~\cite{moghadam2023morphology,liu2022dolce,akrout2023diffusionbased}; \\ DALL$\cdot$E 2~\cite{sagers2022improving}]
        ]
        [Data Efficiency
            [\emph{General FM-based:} ChatGPT and GPT-4~\cite{wu2023exploring}; \\ CLIP~\cite{zhang2023textguided}; GLIP~\cite{yi2023general}; REALM~\cite{zhu2024realm}; BioRAG~\cite{wang2024biorag}]
            [\emph{Medical FM-based:} BrainSegFounder~\cite{cox2024brainsegfounder} \\ REMEDIS~\cite{azizi2023robust}; Biomedical LMs~\cite{mishra2023improving,zhang2023textguided}; \\ Dippel et al.~\cite{dippel2024ai}]
        ]
        [{Large-Scale} \\ {Data Curation}
            [\emph{Datasets for healthcare FMs:} PMC-15M~\cite{zhang2023large}; \\ Virchow~\cite{vorontsove2024clinicalgrade_r08}; OpenPath~\cite{huang2023plip}; Quilt-1M~\cite{ikezogwo2023quilt-1m}; \\ PMC-OA~\cite{lin2023pmc-clip}; HealthCareMagic-100k~\cite{yunxiang2023chatdoctor}]
        ]
    ]
    [Annotation \\ (Section~\ref{section:annotation})
        [Text Annotation
            [Gilson et al.~\cite{gilson2022does}; PathAsst~\cite{sun2023pathasst}; DeID-GPT~\cite{liu2023deid}; \\ Med-PaLM 2~\cite{singhal2023medpalm2}; MedGraphRAG~\cite{wu2024medical}; \\ Tang et al.~\cite{tang2023terminology}]
        ]
        [Image Annotation
            [\emph{Zero-shot SAM:} SAM~\cite{kirillov2023sam}; Hu et al.~\cite{hu2023sam-tumor}; \\ He et al.~\cite{he2023accuracy}; Mazurowski et al.~\cite{mazurowski2023segment}; \\ $\mathrm{SAM^{Med}}$~\cite{wang2023sammed}]
            [\emph{Adapting SAM:} SAM features~\cite{hu2023efficiently}; Poly-SAM~\cite{li2023polyp}; \\ 3DSAM-adapter~\cite{gong20243dsam}; Med SAM Adapter~\cite{wu2023medical}; \\ SAMAug~\cite{zhang2023input}; MedSAM~\cite{ma2024segment}; COSMOS 553K~\cite{huang2023segment}]
        ]
    ]
    [Privacy \\ (Section~\ref{section:privacy})
        [Data Synthesizing
            [SoK~\cite{hu2023sok}; Shibata et al.~\cite{shibata2023practical}; Tang et al.~\cite{tang2023does}; \\ Naresh et al.~\cite{naresh2023privacy}; Dong et al.~\cite{dong2022privacy}; Sei et al.~\cite{sei2024local}]
        ]
    ]
]
\end{forest}
}
\caption{An overview of healthcare data challenges and foundation model-based approaches mentioned in this survey paper.}
\label{fig:issues}
\end{figure}

%% file: content_files/2-FM.tex
\section{Foundation Models}
\label{section:fm}

Foundation models (FMs) are trained on the excessive-scale, wide-ranging data records towards high-level performance on downstream tasks~\cite{bommasani2021foundation}.
The key differentiation of general FM to classic deep learning models is on the \emph{scale} of model size and training data.
First, the success of FMs is built upon the Transformer-style model architecture~\cite{vaswani2017attention}, which can integrate large amounts of information through parallel computing and self-attention mechanisms~\cite{zhou2023comprehensive}.
Second, FM training data contents normally encompass Internet-scale, multi-modal information with labeled and unlabeled annotations.
With the information-rich pre-training, FMs exhibit a comprehensive understanding of concepts and their interrelationships, facilitating the application of knowledge to downstream tasks through \emph{adaptation}.

As a concrete example of FM, large language models (LLMs) are pre-trained on Internet-scale corpora, showing impressive semantic understanding and high-quality text generation~\cite{brown2020gpt3,touvron2023llama,openai2023gpt4}. Recent work also explores reinforcement learning to incentivize reasoning in LLMs~\cite{guod2025deepseekr1_r05}.
Comprehensive surveys summarize the evolving capabilities, limitations, and evaluation considerations of LLMs~\cite{zhaowx2023surveylanguage_r06}.
Open-source LLM families are also rapidly progressing, providing practical backbones for domain adaptation and private deployment (e.g., Qwen2.5 and Qwen3)~\cite{any2024qwen25_r33,yanga2025qwen3technical_r34}.
In particular, LLMs emphasize the engineering design of input context, also known as the \emph{prompt}. This design opens the door for human-machine interaction, allowing us to feed various prompt queries into a well-trained FM towards generating desired content or outcomes.

In this section, we discuss the essential concepts and useful capabilities in the cycle of general FM development, including large-scale pre-training (Section~\ref{sec:fm-pt}), fine-tuning (Section~\ref{sec:fm-ft}), and in-context learning (Section~\ref{sec:fm-icl}). These fundamental techniques are valuable for building a healthcare-focused FM.

\subsection{Large-Scale Model Pre-Training}
\label{sec:fm-pt}

Large-scale model pre-training is an essential approach to building an FM from scratch.
We discuss several core components, including model size, data scale, and self-supervised learning methods, which play vital roles in developing pre-training techniques for building FMs.

\subsubsection{Power of scale}
The Transformer architecture serves as the backbone of FM pre-training, enabling efficient large-scale optimization with self-attention~\cite{vaswani2017attention}. Across NLP and vision, increasing model size, context length, and pre-training corpus scale consistently improves transfer capability and robustness in downstream tasks~\cite{zhou2023comprehensive,dingj2023longnettransformers_r26,liu2025smooth,chowdhery2022palm}. Representative milestones span language scaling and multimodal scaling, including CLIP-style alignment and SAM-style universal segmentation priors~\cite{radford2021clip,kirillov2023sam}.
Recent multimodal pre-training paradigms span contrastive alignment, grounded learning, and unified generation-oriented objectives~\cite{jia2021scaling,singh2022flava,li2022grounded}. In healthcare, related studies report adaptation pipelines that combine domain corpora with deployment-oriented optimization~\cite{gu2023biomedjourney}.
Using document-level corpora allows models to capture long-range dependencies from contiguous text sequences.
As the input context length grows, sequence-length scaling becomes increasingly relevant for long-form clinical notes and longitudinal records~\cite{dingj2023longnettransformers_r26,liu2025smooth}.
For instance, Generative Pre-trained Transformer (GPT)~\cite{radford2018gpt} is a remarkable language model with 117M parameters pre-trained on more than 7,000 unique books (800M words) with various genres from the BooksCorpus dataset.
Bidirectional Encoder Representations from Transformers (BERT)~\cite{devlin2018bert} has 110M parameters for the base model and 340M for the large model. They include BooksCorpus and English Wikipedia (2,500M words) as the pre-training corpus.
With larger training corpora, language model size has expanded from millions to billions of parameters~\cite{chowdhery2022palm}, alongside architectural and optimization improvements.
For example, Text-to-Text Transfer Transformer (T5)~\cite{raffel2020t5} with 11B parameters and 750GB pre-training data converts all NLP tasks into a text-to-text format, enhancing generalization.
GPT-3~\cite{brown2020gpt3} contains 175B parameters and is pre-trained on around 500B word tokens (i.e., fundamental units of text), demonstrating few-shot and zero-shot capabilities.
Pathways Language Model (PaLM)~\cite{chowdhery2022palm} pre-trained on 780B tokens scales up to 540B parameters by employing the Pathways system for efficient distributed training.
A similar scaling pattern appears in modalities beyond text.
The DALL$\cdot$E~\cite{ramesh2021dalle} model with 12B parameters learns rich visual representations on 250M text-image pairs collected from the Internet.
CLIP~\cite{radford2021clip} is a vision-language FM with 304M parameters pre-trained on 400M text-image pairs.
The Segment Anything Model (SAM)~\cite{kirillov2023sam} with 632M parameters is trained on SA-1B, a large dataset for image segmentation with over 1B masks on 11M images whose collection is assisted by SAM.
Recent vision backbones and pre-training objectives, such as BEiT~\cite{wangw2023imageforeign_r30}, further improve scalability for high-resolution visual understanding.
Detailed model-by-model chronology, scale statistics, and additional cross-modal examples are moved to \appref{appendix:ext-fm-basics}.

\subsubsection{Self-supervised learning}
As the data scales up, the burden of human annotation for supervised model training becomes a practical challenge. Self-supervised learning (SSL) is widely used for FM pre-training without the need for labeled data~\cite{devlin2018bert,chen2021mocov3,radford2021clip}. We categorize SSL research into two major approaches: input reconstruction and contrastive learning.

\emph{Input reconstruction} learns by recovering masked or missing content (e.g., masked language/image modeling), while \emph{contrastive learning} aligns semantically related views and separates irrelevant pairs in representation space~\cite{devlin2018bert,he2022mae,xie2022simmim,chen2020simple}. Recent pipelines increasingly combine both objectives for stronger generalization~\cite{chen2021mocov3,radford2021clip,oquab2023dinov2,simeoni2025dinov3}.
For instance, BERT~\cite{devlin2018bert} is pre-trained to perform masked language modeling, where a proportion of tokens are masked from the input sequence, and the model is purposely trained to predict the masked tokens.
Similarly, masked image modeling is a common task for learning visual representations~\cite{he2022mae,xie2022simmim}, where image patches are masked during model training and then the model is required to reconstruct the masked patches.
In NLP, autoregressive language modeling (e.g., GPT~\cite{radford2018gpt}) predicts the next token conditioned on the previous context, complementing masked language modeling.
SimCLR~\cite{chen2020simple} and MoCo-v3~\cite{chen2021mocov3} are examples of applying two random data augmentations to each training image to generate positive samples and recognize different images as negative samples to train a vision transformer.
Besides, Contrastive Language-Image Pre-training (CLIP)~\cite{radford2021clip} views image-text pairs as positive samples and irrelevant image and text as negative samples to learn aligned visual and text representations.
Recent methods increasingly combine input reconstruction and contrastive learning. For instance, DINOv2~\cite{oquab2023dinov2} (and follow-up variants, e.g., DINOv3~\cite{simeoni2025dinov3}) trains a student network to match image representations with a slowly updating teacher network in a contrastive manner, where the two networks take different crops of the same image as input. The student network is required to reconstruct masked image patches.
For example, PubMedBERT~\cite{gu2021pubmedbert} is a self-supervised BERT model pre-trained on 14M PubMed abstracts from scratch without manual annotation, achieving state-of-the-art results on various biomedical NLP tasks.
MoCo-CXR~\cite{sowrirajan2021moco-cxr} pre-trains a visual model on abundant unlabeled chest X-ray images and reports strong representations that can outperform supervised baselines in downstream settings.
In healthcare, SSL pre-training on unlabeled corpora improves data-efficient adaptation for biomedical NLP and medical imaging~\cite{krishnan2022self,gu2021pubmedbert,sowrirajan2021moco-cxr}. Additional method-level comparisons are provided in \appref{appendix:ext-fm-basics}.

\input{content_files/figures/medical-fm}
Figure~\ref{fig:medical-fm} summarizes the role of foundation models across pre-training, adaptation, and downstream healthcare tasks.

\subsection{Fine-Tuning}
\label{sec:fm-ft}

The paradigm of pre-training and fine-tuning has dominated deep learning since self-supervised learning enabled large-scale model pre-training.
In detail, fine-tuning refers to updating pre-trained model parameters through forward and backward passes and gradient descent on task-specific supervised data.
Fine-tuning is a critical technique in constructing medical domain-specific FMs~\cite{gema2023peft-llama,li2023llava-med,saab2024med-gemini}.
However, conventional fine-tuning poses significant challenges, including high cost, extensive data requirements, and limited generalization to unseen tasks.
By contrast, we introduce two essential techniques beyond basic fine-tuning to address those issues: parameter-efficient fine-tuning and instruction tuning.

\subsubsection{Parameter-efficient fine-tuning}
Parameter-efficient fine-tuning (PEFT) is a family of fine-tuning approaches that only updates a small set of model parameters while keeping most of the pre-trained weights fixed.
These methods alleviate critical challenges caused by updating all parameters of a large FM during full fine-tuning, such as the excessively high cost, overfitting on small downstream datasets, and the risk of catastrophic forgetting~\cite{ding2023parameter}.
PEFT approaches excel in scenarios with limited data availability and effectively retain the valuable knowledge acquired through FM pre-training.

We discuss essential PEFT methods.
Representative PEFT families include bias-only tuning, adapter modules, prompt-based tuning, and low-rank updates (LoRA)~\cite{zaken2021bitfit,houlsby2019adapter,hu2021lora}. These methods improve adaptation efficiency with reduced trainable parameters, and have shown practical value in healthcare settings~\cite{gema2023peft-llama}. Detailed algorithm-level comparisons are moved to \appref{appendix:ext-fm-basics} and \appref{appendix:ext-medfm}.
Representative PEFT algorithms also include BitFit, adapter tuning, prompt tuning, prefix tuning, and LoRA variants for different parameter and compute budgets~\cite{zaken2021bitfit,houlsby2019adapter,lester2021prompt-tuning,li2021prefix-tuning,hu2021lora}.
Recent studies discuss matching PEFT configuration to clinical data regime and task structure, and explore domain-aware adapter or prompt design under low-resource and cross-site conditions~\cite{ding2023parameter,lu2024pathotune}.

\subsubsection{Instruction tuning}
Instruction is defined as the linguistic description of a task along with its corresponding task-specific data sample.
Instruction tuning refers to fine-tuning FMs on supervised instruction datasets with LLMs, helping to understand the instruction~\cite{ouyang2022training}.
This method enhances zero-shot performance on new tasks and improves the generalization capability of the fine-tuned FM.
The outcome of instruction tuning is influenced by both the proficiency of the pre-trained FM and the quality of the instruction-following data~\cite{alpaca}.
For instance, Fine-tuned LAnguage Net (FLAN)~\cite{wei2021flan} is an LLM fine-tuned on tens of NLP datasets via natural language instructions, outperforming GPT-3 on most of the zero-shot evaluation datasets.
In addition, InstructGPT~\cite{ouyang2022training} and ChatGPT~\cite{chatgpt} are LLMs fine-tuned on instructions with human feedback to promote user alignment, resulting in chatbots with more truthful outputs.
Alpaca~\cite{alpaca} is a LLaMA model~\cite{touvron2023llama} fine-tuned on 52K InstructGPT-generated instructions, achieving performance similar to InstructGPT but with a significantly smaller model size. 
Instruction tuning has also been applied to build healthcare FMs~\cite{li2023llava-med,yunxiang2023chatdoctor,chen2025cost}.

\subsection{In-Context Learning}
\label{sec:fm-icl}

Large language models (LLMs), as a concrete example of FM, bring a promising learning paradigm termed in-context learning (ICL), also known as prompt engineering~\cite{brown2020gpt3}.
\emph{Prompt} essentially represents an input query that guides the model's output, which has proven to greatly impact the model's performance even without the need for fine-tuning. By merely adjusting the input query of LLMs during inference instead of updating any parameters of LLMs, we can flexibly prompt LLMs to generate desired outputs without fine-tuning them on downstream tasks~\cite{brown2020gpt3,dong2022survey-icl}.

ICL promises to extend powerful LLMs into downstream tasks by injecting contextual information.
Mainstream ICL strategies include zero/few-shot prompting and reasoning-oriented variants (e.g., chain-of-thought, self-consistency, tree-of-thought, and self-refinement)~\cite{kojima2022cot,wang2022cot-sc,yao2023tot,madaan2023self-refine}. Detailed prompting examples are provided in \appref{appendix:ext-fm-basics}.
In healthcare-oriented deployments, these prompting strategies are often combined with retrieval cues to stabilize reasoning traces and improve response grounding in domain-specific contexts~\cite{lewis2020retrieval,zhao2024retrieval,wu2024medical}.

ICL can provide grounded knowledge to compensate for the lag in pre-training data of FMs and the lack of domain-specific knowledge.
\emph{Retrieval-augmented generation} (RAG) is a technique that leverages external knowledge retrieval mechanisms to gather additional relevant information for ICL, enhancing the generation quality of FMs \cite{lewis2020retrieval,zhao2024retrieval}.
In RAG, FMs (typically LLMs) first generate queries from user input, which are then used to retrieve information from a knowledge base or external documents.
The retrieved context is integrated with the input, allowing FMs to produce more informed and accurate outputs.

FMs start to demonstrate generalization power in the medical field using ICL techniques~\cite{roy2023sam-md,lyu2023gpt-translate-radiology}.
For instance, ChatGPT can effectively translate radiology reports into plain language by using well-designed prompts that inform the model about the structure of the report and contents of each paragraph~\cite{lyu2023gpt-translate-radiology}.
Also, the general-purpose SAM performs well on abdominal CT organ segmentation when provided point and bounding box prompts in an oracle manner~\cite{roy2023sam-md,zhu2025guiding,huang2024cat}.
Besides, RAG methods show great potential in enhancing the capabilities of LLMs for medical-related NLP tasks \cite{zhu2024realm,wu2024medical}.

%% file: content_files/figures/medical-fm.tex
\begin{figure}[t]
    \centering
    \includegraphics[width=\linewidth]{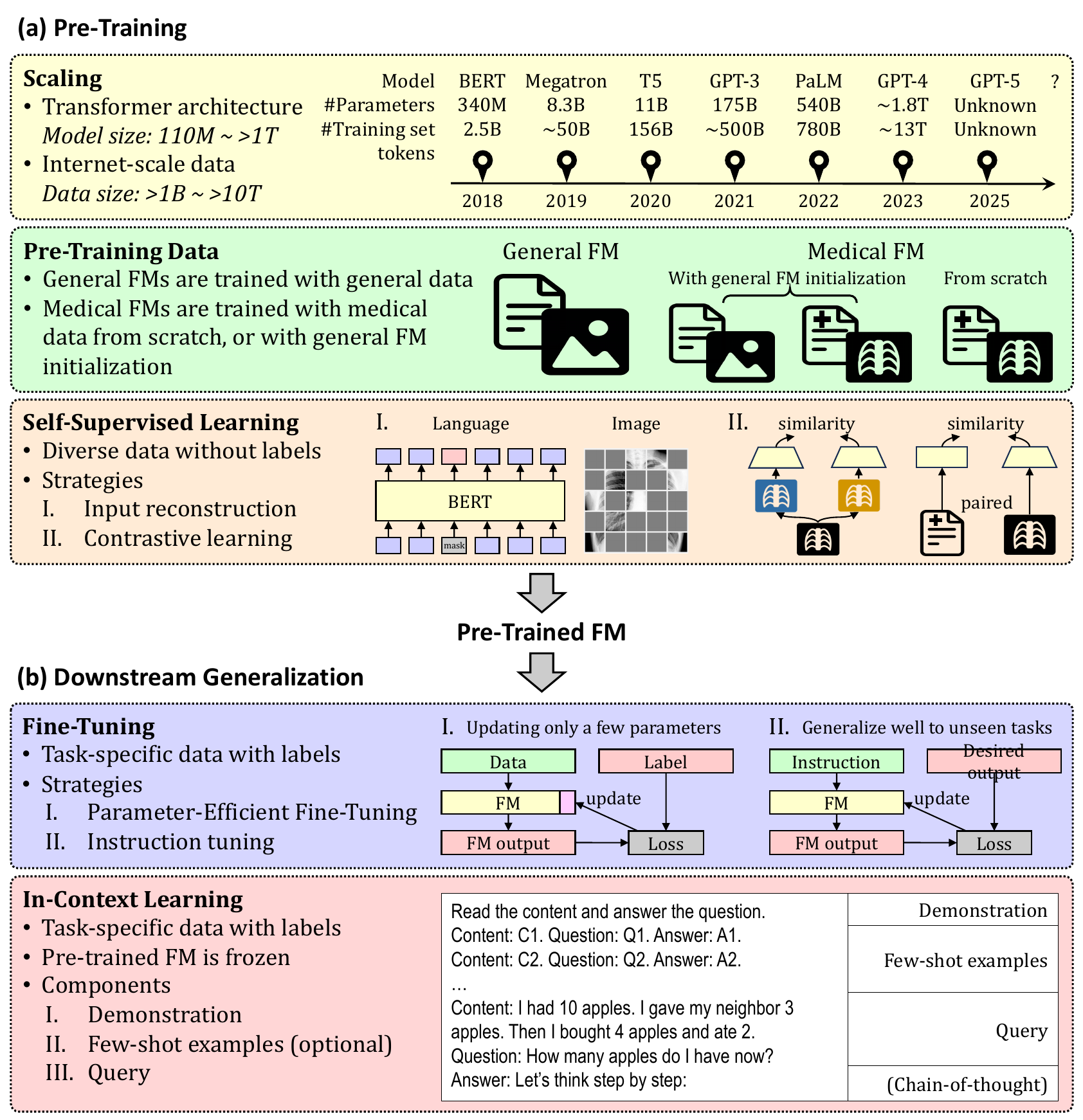}
    \caption{Foundation model (FM) in healthcare.}
    \label{fig:medical-fm}
\end{figure}

%% file: content_files/3-HealthcareFM.tex
\section{Foundation Models in Healthcare}
\label{section:medfm}

The growth of FM analytics offers insights into healthcare applications~\cite{qiu2023survey-ai-health,zhang2023challenges,dengfoundation}. We review key techniques, tools, and applications addressing multiple aspects of FM in healthcare. We exhibit how general-purpose FMs can be applied in the healthcare field (Section~\ref{sec:medfm-gfm}). We present medical-focused FMs and demonstrate pre-training benefits gained from general FMs (Section~\ref{sec:medfm-mfm}).

\subsection{Adapting General Foundation Models in Medicine and Healthcare}
\label{sec:medfm-gfm}

Research efforts have started to assess FM's superior capability in the medical domain~\cite{nori2023gpt4-on-medical,gema2023peft-llama,ma2024segment}. In these studies, we identify two core techniques, including parameter-efficient fine-tuning (PEFT) and in-context learning (ICL).

\subsubsection{Adapting via parameter-efficient fine-tuning (PEFT)}
PEFT methods have been applied to adapt FMs to medical tasks.
For instance, Dutt et al.~\cite{dutt2023parameter} demonstrate that PEFT methods significantly outperform full fine-tuning of FMs in data-limited scenarios for medical image classification and text-to-image generation tasks.
Gema et al.~\cite{gema2023peft-llama} propose a two-stage PEFT framework to adapt LLaMA~\cite{touvron2023llama} to a broad range of clinical tasks. In this work, the first stage applies LoRA~\cite{hu2021lora} to fine-tune LLaMA on clinical notes, building Clinical LLaMA-LoRA, a clinical FM; the second stage again applies LoRA to adapt the clinical FM to downstream tasks.
Similarly, Van Veen et al.~\cite{van2023radadapt} apply LoRA to fine-tune T5 models~\cite{raffel2020t5,lehman2023clinical-t5} for radiology report summarization.
They also apply LoRA together with in-context learning for clinical text summarization tasks, showing improved performance over human experts~\cite{vanveen2023clinical}.

Recent studies further compare LoRA, prompt-based tuning, and fairness-aware PEFT selection under low-data clinical settings~\cite{gema2023peft-llama,zu2024embedded,dutt2023fairtune,ding2023parameter}. We summarize detailed PEFT comparison evidence in \appref{appendix:ext-medfm}.

\subsubsection{Adapting via in-context learning (ICL)}
ICL has proven to be effective in adapting FMs, especially large language models (LLMs), to a variety of healthcare tasks. With carefully designed task-specific input context (i.e., prompts), the FM can perform well on healthcare tasks without modifying any model parameters.
For instance, Nori et al.~\cite{nori2023gpt4-on-medical} evaluate GPT-4~\cite{openai2023gpt4} on the United States Medical Licensing Examination (USMLE) without specially crafted prompts. GPT-4 shows its promising zero-shot performance without adding relevant medical context data.
Lyu et al.~\cite{lyu2023gpt-translate-radiology} leverage ChatGPT~\cite{chatgpt} to translate radiology reports into plain language for report understanding and translation. The experiments show that by using a clearer and more structured prompt, the overall translation quality can be increased.
Deng et al.~\cite{deng2025segment} evaluate the zero-shot performance of SAM on tumor segmentation, non-tumor tissue segmentation, and cell nuclei segmentation on whole slide images (WSI), demonstrating that SAM performs well on large connected objects on pathological scans.
Chen et al. propose Diagnosis of Thought (DoT) prompting~\cite{chen2023dot} to assist professionals with cognitive distortion detection. DoT diagnoses mental illness by prompting LLMs to sequentially perform subjectivity assessment, contrastive reasoning, and schema analysis.

\subsection{Pre-Training Healthcare Foundation Models}
\label{sec:medfm-mfm}

Researchers make efforts to pre-train FMs based on large-scale unlabeled healthcare data for health record examination~\cite{gu2021pubmedbert,alsentzer2019clinical-bert,singhal2022med-palm}, medical imaging diagnosis~\cite{azizi2023robust,huang2023plip}, and protein sequence analysis \cite{lin2023esm,cheng2023alphamissense}.
In principle, the pre-training process can be summarized into two major aspects: pre-training strategy and model initialization.

\subsubsection{Pre-training strategy}
Healthcare FM pre-training typically utilizes a range of pre-training strategies derived from general-domain FMs due to their potential generalization power.

One pre-training strategy is masked language/image modeling, following BERT~\cite{devlin2018bert} and masked autoencoder (MAE)~\cite{he2022mae}.
For instance, SciBERT~\cite{Beltagy2019SciBERT} and PubMedBERT~\cite{gu2021pubmedbert} are pre-trained on multi-domain scientific publications and biomedical domain-specific corpora, respectively, based on the BERT strategy.
Complementary to masked modeling, autoregressive pre-training supports generative biomedical LMs. BioGPT~\cite{luo2022biogpt} is pre-trained on PubMed\footnote{\url{https://pubmed.ncbi.nlm.nih.gov/}} abstracts following GPT-2~\cite{radford2019gpt2} for generative language tasks.
RETFound~\cite{zhou2023retfound} is an FM for retinal image disease detection, pre-trained on a large collection of unannotated retinal images to reconstruct input images with 75\% masked patches, following MAE.
Similarly, General Expression Transformer (GET)~\cite{fu2023get} is an FM for modeling transcriptional regulation across 213 human cell types. GET is pre-trained to predict the motif binding scores of masked regulatory elements in the input to learn regulatory patterns.

Contrastive learning is another important pre-training strategy for medical FMs, including medical vision pre-training and vision-language alignment with domain-specific image-text pairs~\cite{azizi2023robust,huang2023plip,lu2024visual,kim2024transparent}. Additional model-by-model examples are provided in \appref{appendix:ext-medfm}.

Masked modeling is well-suited for dense prediction tasks, including text generation and image segmentation, making it widely used in healthcare FM pre-training. In contrast, contrastive learning is more effective for representation learning and multi-modal understanding, offering a broader generalization but requiring a larger scale of data for negative sampling. Researchers have combined masked modeling and contrastive learning to capture representations at different scales in pathological imaging. For example, Prov-GigaPath~\cite{xu2024whole} is a whole-slide FM for digital pathology that employs a tile-level encoder trained following DINOv2 and a slide-level encoder trained via MAE.

Beyond these pre-training strategies, inspiration has been drawn from language models' ability to process sequential inputs, leading to large pre-trained models for protein sequence tasks.
ESM-2~\cite{lin2023esm} is a Transformer model with 15B parameters pre-trained on millions of protein sequences to predict protein structure directly from amino acid sequences, fast and accurately. ESM-2 illustrates the immense potential of LLMs to learn patterns in protein sequences across evolution.
AlphaMissense~\cite{cheng2023alphamissense} pre-trains an AlphaFold-like model~\cite{jumper2021alphafold} to predict protein structure via protein language modeling. It then fine-tunes the model with an additional variant pathogenicity classification objective on human and primate variant population frequency databases. AlphaMissense achieves state-of-the-art performance on missense variant pathogenicity prediction.

\subsubsection{Model initialization}
Healthcare FM pre-training benefits from utilizing general FMs as the initial model to leverage their massive information.
With proper model initialization, we recognize that much less data and fewer training epochs are needed for domain-specific pre-training of medical FMs.
Representative initialization pathways include biomedical language adaptation from BERT/LLaMA families and vision-language transfer from CLIP-style pre-training~\cite{lee2020biobert,wu2023pmc-llama,singhal2022med-palm,eslami2021pubmedclip}. We defer additional multilingual and task-specific initialization examples to \appref{appendix:ext-medfm}. Medical data heterogeneity has long been an obstacle to FM pre-training, generalization, and evaluation~\cite{huang2020fusion,rajpurkar2022ai-health}. This challenge is further compounded by privacy and governance constraints that limit cross-site data sharing~\cite{hathaliya2020security-privacy}.

%% file: content_files/4-modality.tex
\section{Multi-Modal Data Fusion}
\label{section:modality}

Data fusion is a useful strategy for aggregating the information of various medical data modalities towards improved decision-making in healthcare. Standard fusion methods can be conceptually categorized into three types~\cite{stahlschmidt2022multimodal}: early fusion, joint fusion, and late fusion.
In early fusion, data from different modalities are combined at the input and passed to the subsequent network~\cite{zhao2021deepomix,fu2020gene}.
Joint fusion processes data from each modality through independent networks before fusing their representative feature maps as the input of the subsequent network~\cite{braman2021deep,kawahara2018seven}.
Late fusion processes data from each modality through individual networks and fuses their output to make the final outcomes~\cite{reda2018deep,qiu2018fusion}.
However, conventional methods suffer from a lack of scalability, generalizability, and cross-modal understanding~\cite{huang2020fusion}. For example, early and joint fusion methods often require sufficient task-specific training data and computational resources, while late fusion methods usually show weaknesses in multi-modal information integration~\cite{budikova2017fusion}.
Figure~\ref{fig:modality} illustrates the healthcare data modalities, data fusion strategies, and associated healthcare tasks.

\input{content_files/figures/modalities.tex}

\subsection{Multi-Modal Healthcare Data}

Healthcare data is inherently multi-modal, encompassing structured electronic health records (EHRs), unstructured clinical notes, medical images, genomic sequences, and wearable sensor outputs.
Each modality presents unique characteristics associated with diseases and patient outcomes, necessitating specialized techniques to address the data heterogeneity and leverage their collective potential.

\emph{Structured EHR data} provides organized, tabular information but often lacks contextual depth, requiring tailored model architectures to fully capture its information~\cite{guo2024multi,wornow2023shaky}.
CLMBR-T-base~\cite{wornow2023ehrshot} processes medical codes as tokens by using an autoregressive language model pre-trained on full longitudinal structured EHRs.
BEHRT~\cite{li2020behrt} is represented as a BERT-like model and concatenates patient procedures and diagnoses into long sequences, while ExBEHRT~\cite{rupp2023exbehrt} stacks them vertically to mitigate input length constraints.
EHRMamba~\cite{fallahpour2024ehrmamba} leverages the Mamba architecture~\cite{gu2023mamba} to efficiently process long EHR sequences, addressing challenges in handling extensive medical histories.
Furthermore, state space models like Mamba are being adapted as vision foundation models to efficiently process high-resolution whole slide images and volumetric scans~\cite{lu2025graph,liu2024swin}.

\emph{Unstructured clinical notes} are rich in narrative detail and require advanced NLP techniques to extract meaningful insights~\cite{vanveen2023clinical,alsentzer2019clinical-bert}.
Large language models (LLMs) pre-trained on clinical data have demonstrated effectiveness in handling clinical tasks.
Clinical BERT~\cite{alsentzer2019clinical-bert} is one of the first LLMs pre-trained on MIMIC-III~\cite{johnson2016mimic-iii} notes.
Med-PaLM 2~\cite{singhal2023medpalm2} enhances generalization across diverse clinical tasks with a larger model size and broader data scale.

\emph{Medical images} are high-dimensional and heterogeneous, spanning 2D radiographs and fundus photos, 3D CT/MRI volumes, ultrasound videos, and gigapixel pathology slides.
They often exhibit substantial site- and protocol-dependent shifts, which complicates harmonization and downstream model transfer~\cite{ma2024segment,azizi2023robust,chen2023uni}.

Multi-modal FMs address these data challenges through architectures designed for heterogeneous clinical data fusion~\cite{eslami2021pubmedclip,moor2023med-flamingo,saab2024med-gemini}. Key techniques such as cross-attention mechanisms~\cite{alayrac2022flamingo}, hierarchical encoders~\cite{ge2024convllava}, and shared latent representations~\cite{radford2021clip} are helpful to integrate information across clinical data modalities. We discuss multi-modal FM techniques from two perspectives. First, a joint-modal pre-training can enhance data fusion at scale, improving transferability to downstream healthcare tasks. Second, LLMs possess strong comprehension and reasoning abilities, facilitating the cross-modality interaction by processing aggregated multi-modal inputs.

\subsection{Data Fusing via Joint-Modal Pre-Training}

FMs can handle multiple modalities via pre-training on massive-scale paired multi-modal data in a joint-modal mode to obtain a high-level understanding of inter-modality relationships.
Representative pre-training families align image-text features in shared latent spaces and then transfer this alignment to healthcare domains for robust zero-/few-shot adaptation~\cite{radford2021clip,eslami2021pubmedclip,zhang2022contrastive}. Recent medical work further scales joint-modal pre-training from CXR-report pairs to CT/MRI and pathology-report settings. Detailed model-by-model inventories and modality-specific variants are moved to \appref{appendix:ext-modality}.

Beyond early vision-language pre-training (VLP), joint-modal and self-supervised pre-training has rapidly expanded to clinical-grade foundations across pathology, ophthalmology, dermatology, radiology, ultrasound, and dentistry. A shared data-centric trend is the dependence on curated real-world corpora, cross-site validation, and more diverse downstream benchmarks to quantify robustness and distribution shifts~\cite{chen2023uni,developmentandvalidation2024_0437bee7,echofmfoundationmodel2025_45c40046}. We move the full domain-wise case inventory and citation clusters to \appref{appendix:ext-modality}.

\subsection{Data Fusing via LLMs}

Transformer-style LLMs possess powerful semantic understanding capability via the attention mechanism~\cite{brown2020gpt3}, which can be transferred to multi-modal settings.
To be specific, data from different modalities can be aggregated as the prompt input of an LLM (i.e., a sequence of tokens). These combined multi-modal inputs are then fused through the Transformer blocks in the LLM, exchanging information via attention layers~\cite{wang2023visionllm,liu2023vinst}.
Conceptually, LLM-driven fusion can be viewed as (i) token-level multimodal prompting that conditions an LLM on modality-specific encoders, (ii) instruction-tuned multimodal chat models that align encoders and dialog behavior, and (iii) tool-augmented agents that ground generation with external knowledge and specialized clinical models.
Recent surveys use the term \emph{medical agents} to describe stateful, tool-driven systems that act over multimodal, longitudinal clinical data and coordinate tool calls within clinical workflows~\cite{hu2025medicalagentsurvey,zhu2025menti,huang2024tool}.
Clinically, \emph{copilots} keep clinicians in the loop for interpretation and documentation, whereas \emph{agents} introduce state, memory, and tool use to execute multi-step plans with explicit constraints.
From a data-centric perspective, this shift makes longitudinal context, interaction traces, and tool-call logs part of the core data substrate to curate and benchmark, rather than optional metadata~\cite{hu2025medicalagentsurvey,mehandru2024agentsclinic}.
Representative multimodal-LLM designs include cross-attention conditioned language models and connector-based architectures (e.g., Flamingo/BLIP-style and LLaVA-style systems) that align vision encoders with instruction-following language models~\cite{alayrac2022flamingo,li2023blip2,liu2023vinst}. Healthcare adaptations then apply these paradigms to MedVQA, report understanding, and multimodal reasoning workflows~\cite{moor2023med-flamingo,li2023llava-med,nori2023gpt4-on-medical}. Additional cross-modality and domain-specific examples are listed in \appref{appendix:ext-modality}.

In domain-specific settings, multimodal copilots and agentic systems are expanding beyond image-text tasks to oral health, ophthalmology, pathology workflows, biosignals, and 3D neuro/cardiac imaging, which further emphasizes temporal context curation and workflow-level evaluation~\cite{lu2024multimodal,ecglmunderstanding2025_a9510e52,caro2023brainlm}. Detailed modality-specific examples are provided in \appref{appendix:ext-modality}.
We believe that joint-modal pre-training techniques, combined with LLMs' logical reasoning capabilities, could benefit healthcare data fusion of modalities beyond the scope of language and vision.

%% file: content_files/figures/modalities.tex
\begin{figure}[t]
    \centering
    \includegraphics[width=\linewidth]{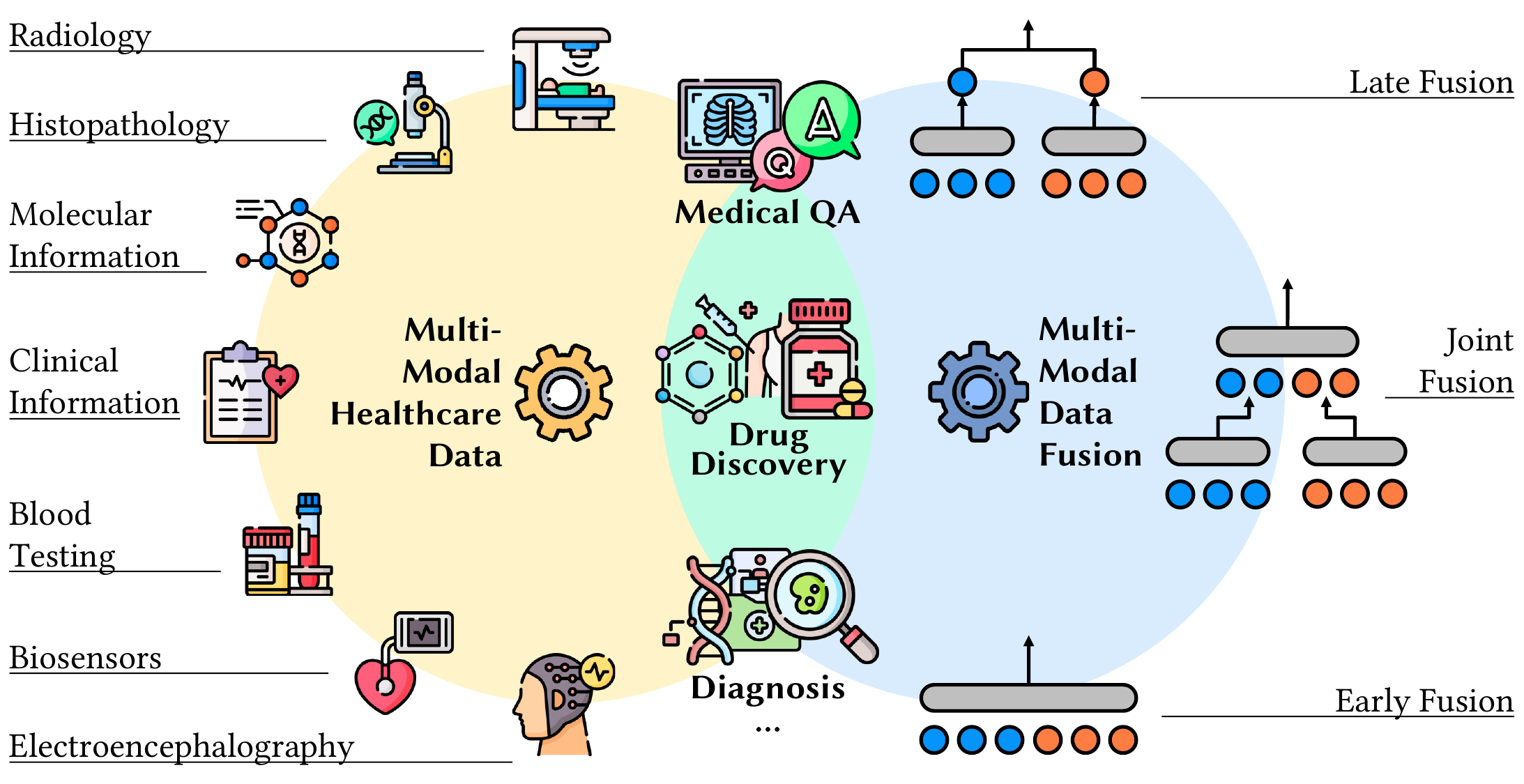}
    \caption{Multi-modal fusion of healthcare data in the FM era. Conventional fusion approaches are enhanced by joint-modal pre-training and comprehensive FMs such as LLMs, enabling downstream applications such as medical QA, drug discovery, and diagnosis.}
    \label{fig:modality}
\end{figure}

%% file: content_files/5-quantity.tex
\section{Data Quantity}
\label{section:quantity}

To apply FMs in the healthcare system safely and responsibly, we must address a variety of data quantity hurdles.
A key issue is the trade-off between the limited information provided by constrained datasets and the essential information required for training a robust model on the specific healthcare task.
Such a disparity can result in inadequate downstream model training, yielding non-robust, inaccurate, and even unreliable outcomes.
However, public patient data records are often scarce due to the rigorous patient privacy protection regulations. In addition, real-world dataset curation, involving data collection, cleaning, and annotation, is costly.
While deep learning has shown its promise, the barrier of limited data access in healthcare remains a significant roadblock for scalable AI-based research.

The emergence of FMs holds the potential to alleviate the data quantity challenges in healthcare applications.
Pre-trained on large-scale datasets consisting of billions of samples, FMs obtain vast amounts of information to compensate for limited data in downstream healthcare tasks.
In this section, we discuss representative approaches to handling data quantity challenges using FMs from the perspectives of data augmentation (Section~\ref{sec:quantity-augmentation}) and data efficiency (Section~\ref{sec:quantity-efficiency}).
We then introduce the curation of large-scale healthcare datasets from the Internet to support healthcare FM pre-training (Section~\ref{sec:quantity-internet}).

\subsection{Data Augmentation}
\label{sec:quantity-augmentation}

Data augmentation is a common strategy in machine learning for addressing the issue of data sample limitation~\cite{shorten2019survey,shorten2021text}.
Conventional augmentation techniques include resizing, clipping, and flipping for images~\cite{krizhevsky2012imagenet,szegedy2016rethinking} and synonym replacement, random insertion, and back translation for text~\cite{sennrich2015improving,wei2019eda}.
Data augmentation has also been applied within the healthcare domain~\cite{kang2023benchmarking,tellez2019quantifying,shen2022randstainna}.
Yet, these techniques only manipulate existing data samples and maintain limited information entropy since no external information is introduced beyond the existing data distribution.

Information-rich generative FMs have brought a remarkable shift to healthcare data augmentation for both imaging and text~\cite{kazerouni2023diffusion,nori2023gpt4-on-medical}.
Pre-trained with a vast set of knowledge, FMs enable the transfer of general insights to the healthcare domain beyond the scope of limited datasets.

In medical image augmentation, diffusion and vision-language foundation models can generate synthetic data across multiple imaging modalities, while text-oriented FMs can provide synthetic instruction and clinical language variants for downstream fine-tuning~\cite{kazerouni2023diffusion,pinaya2022brain,tang2023does}. We move the full model-by-model augmentation case list to \appref{appendix:ext-quantity}.
Reported applications span CXR, CT, MRI, pathology, and dermatology synthesis pipelines, including cycle-consistent diffusion variants for rare-pattern generation~\cite{chambon2022adapting,liu2022dolce,moghadam2023morphology,song2024cycle}.
This direction introduces external information from general-domain visual data.
The effectiveness of diffusion-based data augmentation is notably evident across various imaging modalities, including chest X-rays (CXRs)~\cite{chambon2022adapting}, computed tomography (CT)~\cite{liu2022dolce}, brain magnetic resonance imaging (MRI)~\cite{pinaya2022brain}, histopathology~\cite{moghadam2023morphology}, and dermatology~\cite{akrout2023diffusionbased}.
Despite modality-specific differences, such transfer has been explored across diverse healthcare contexts.
Beyond diffusion, vision-language foundation models have also been explored for generating realistic radiographs as controllable synthetic data for chest imaging~\cite{avisionlanguage2024_502e1fd2}.
For instance, Pinaya et al.~\cite{pinaya2022brain} harness latent diffusion models~\cite{rombach2022high} from general vision to generate synthetic images from high-resolution 3D brain scans.
Furthermore, Sagers et al.~\cite{sagers2022improving} showcase the potential of DALL$\cdot$E 2~\cite{ramesh2022hierarchical}, a text-to-image diffusion model, to generate realistic depictions of skin diseases across varying skin types.
Meanwhile, FMs demonstrate positive performance in clinical text mining through textual data augmentation. Employing ChatGPT~\cite{chatgpt} to generate high-quality synthetic data with labels has proved beneficial for fine-tuning local models for downstream tasks~\cite{tang2023does}. Similarly, PathGen-1.6M provides a large-scale synthetic pathology image--text corpus that can support vision-language pre-training~\cite{sun2024pathgen16m}.
This approach effectively addresses the challenge of data quantity in clinical text mining.
Similarly, Yuan et al.~\cite{yuan2024large} leverage ChatGPT to modify lexical choice and syntax while maintaining semantic meaning, thereby enhancing patient-trial matching data for downstream fine-tuning.
Using ChatGPT to generate synthetic data can reduce part of the burden of extensive data collection.
Recent studies also explore advanced synthesis techniques such as Mamba-enhanced UNets and cycle-consistent diffusion for highly realistic CT and MRI generation~\cite{song2024cycle,zhou2025glfc,song2024cross}.

\subsection{Data Efficiency}
\label{sec:quantity-efficiency}

Data-efficient approaches leverage FMs to reduce the data volume required by downstream tasks, improving efficiency and addressing data quantity challenges in healthcare.
Serving as a bridge to connect massive upstream data and limited downstream data, these methods enable small healthcare datasets to achieve satisfactory results by incorporating knowledge from pre-trained general-domain FMs.
For example, CITE~\cite{zhang2023textguided} explores the adaptation of general vision FMs, such as CLIP~\cite{radford2021clip} and INTERN~\cite{shao2021intern}, to comprehend pathological images, shedding light on the utilization of medical domain-specific text knowledge to enhance data-efficient pathological image classification.
Similarly, Wu et al.~\cite{wu2023exploring} demonstrate that general large language models (LLMs) such as ChatGPT and GPT-4~\cite{openai2023gpt4} exhibit strong capabilities in handling radiology natural language inference (NLI) tasks even with limited data.

General FMs can efficiently integrate external information from public data sources for downstream tasks using retrieval-augmented generation (RAG) techniques.
RAG is a highly effective method for enabling FMs to acquire grounded domain knowledge, which is publicly accessible but not included in model pre-training.
For instance, BioRAG~\cite{wang2024biorag} leverages an LLM to adaptively select knowledge sources from papers, search engines, and biological data hubs and retrieve information for biological question-reasoning.
REALM~\cite{zhu2024realm} extracts comprehensive representations for EHR data by fusing EHR with information retrieved from a medical knowledge graph, without training on task-specific data.

In addition to general FMs, healthcare-focused FMs are increasingly designed for data efficiency in low-resource settings, including transfer-friendly representation learning and continual adaptation~\cite{azizi2023robust,dippel2024ai,mishra2023improving}. More detailed case comparisons are listed in \appref{appendix:ext-quantity}.
In this context, healthcare-focused FMs offer a practical direction for improving data efficiency in downstream medical applications.
Recent pathology and multimodal studies examine joint optimization of representation transfer, prompt/interface design, and domain-specific calibration under sparse-label and distribution-shift settings~\cite{boecking2022making,liu2023clip}.
For example, REMEDIS~\cite{azizi2023robust} combines large-scale supervised transfer learning with self-supervised learning to reduce data requirements.
It achieves comparable performance to fully supervised models in out-of-distribution settings using only 1-33\% of downstream data.
Similarly, Dippel et al.~\cite{dippel2024ai} train an unsupervised anomaly detection FM that outperforms supervised models, especially for rare diseases.
BrainSegFounder~\cite{cox2024brainsegfounder}, a neuroimage segmentation FM pre-trained with generally healthy brain MRI images, shows great potential in reducing the data required for downstream fine-tuning.
As for medical text data, Mishra et al.~\cite{mishra2023improving} report that pre-trained medical text encoders perform well on low-prevalence diseases, supporting their use in addressing data quantity challenges across healthcare domains.
Moreover, Yi et al.~\cite{yi2023general} introduce continual learning (including sequential learning and rehearsal learning) based on medical FMs as a data-efficient learning paradigm for this setting.

\subsection{Large-scale Data Curation}
\label{sec:quantity-internet}

The data quantity challenge in the healthcare system strongly limits the performance of downstream applications. FMs provide a novel means to leverage large-scale data from public web resources and institutional archives. To support healthcare FM pre-training, curating and processing large-scale healthcare datasets is critical.
A representative collection of healthcare data requires high-quality data sources and effective data extraction strategies.

One common approach is to curate multi-source corpora from biomedical repositories, institutional archives, and online platforms, then standardize and filter the data for FM pre-training~\cite{lin2023pmc-clip,vorontsove2024clinicalgrade_r08,ikezogwo2023quilt-1m}. Additional curation examples are provided in \appref{appendix:ext-quantity}.
In practice, many pipelines first extract text and images from PubMed Central (PMC), an Internet-scale archive of biomedical literature, and then combine this source with institution-level collections.
For instance, PMC-OA~\cite{lin2023pmc-clip} and PMC-15M~\cite{zhang2023large} extract image-caption pairs from PMC articles.
Virchow~\cite{vorontsove2024clinicalgrade_r08} collects pathology scans from the Memorial Sloan Kettering Cancer Center (MSK), illustrating institution-scale data curation from a clinical research partner.
OpenPath~\cite{huang2023plip} utilizes pathology hashtags to collect pathology-related tweets with certain keywords or high numbers of likes.
Quilt-1M~\cite{ikezogwo2023quilt-1m} searches YouTube for histopathology videos and obtains corresponding text through speech-to-text techniques and LLM postprocessing.
HealthCareMagic-100k~\cite{yunxiang2023chatdoctor} collects patient-physician dialogues from an online medical consultation platform.

\input{content_files/figures/quantity-annotation.tex}
Figure~\ref{fig:quantity-annotation} illustrates representative FM-enabled strategies for mitigating data quantity limits and annotation burden.

%% file: content_files/figures/quantity-annotation.tex
\begin{figure}[t]
    \centering
    \includegraphics[width=0.82\linewidth]{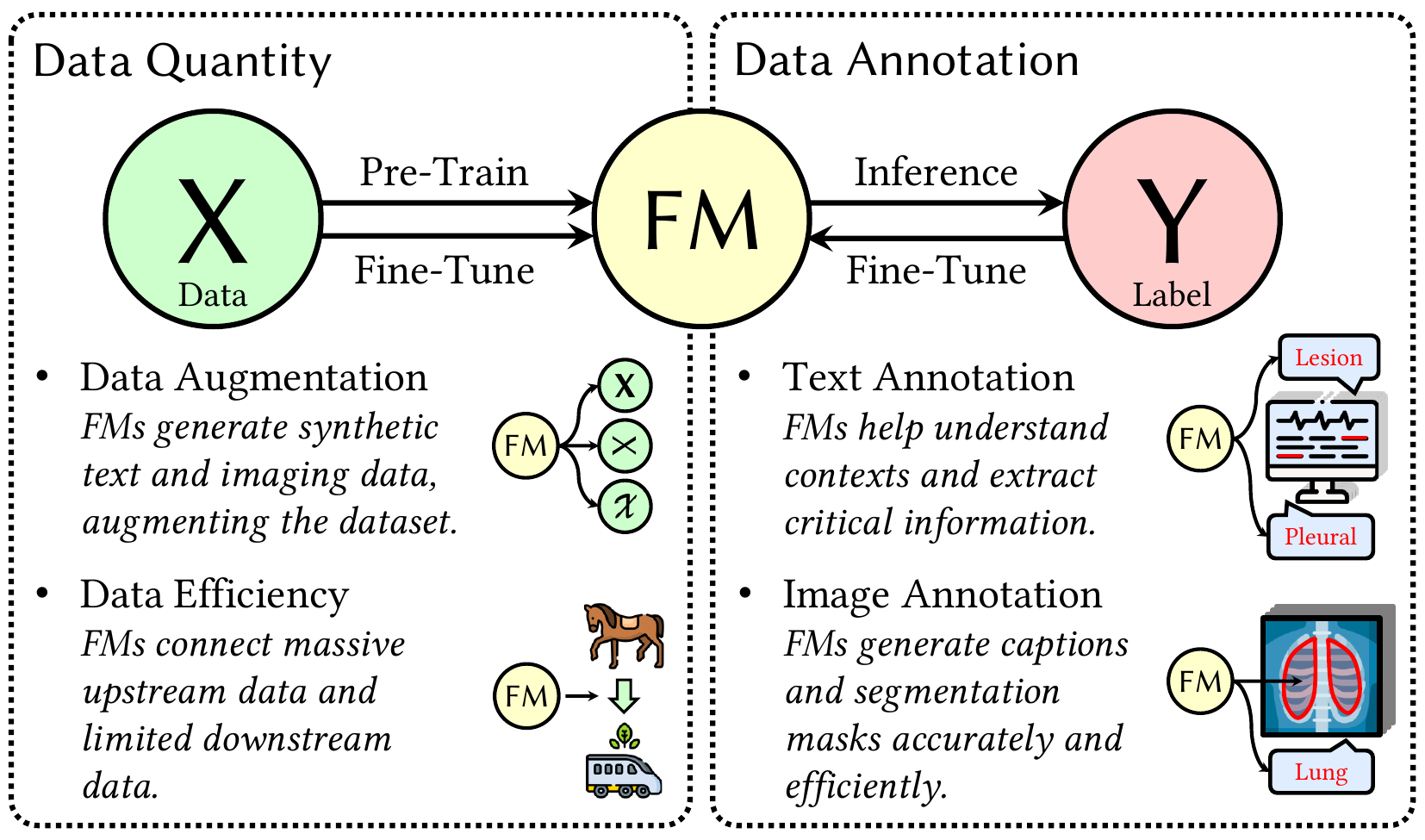}
    \caption{Foundation models address data quantity and data annotation challenges. \textbf{Left:} Foundation models can mitigate data quantity limitation by data augmentation and improved data efficiency. \textbf{Right:} Foundation models can help both healthcare text and medical image annotation.}
    \label{fig:quantity-annotation}
\end{figure}

%% file: content_files/6-annotation.tex
\section{Data Annotation}
\label{section:annotation}

Data annotation is a critical step in computational healthcare, involving the labeling of medical data such as medical images, electronic health records, and genomic sequences~\cite{belle2015big}.
By assigning informative metadata or class labels, data annotation enriches the raw dataset with a nuanced layer of human expertise and contextual understanding, providing valuable insights for healthcare education, diagnostics, and AI applications~\cite{lee2017big,willemink2020preparing}.
As such, data annotation plays a pivotal role in leveraging FMs to diagnose diseases, personalize treatments, and advance a more informed and efficient healthcare system~\cite{razzak2018deep}.
However, numerous challenges persist in data annotation, including the shortage of professional annotators and the complexity of the annotation process~\cite{dgani2018training}.

The scalability of FMs offers a promising solution to the labor-intensive nature of healthcare data annotation. The FM-based scalable annotation paradigm is developing rapidly in the medical field for the following reasons.
First, core FM techniques such as tokenization and modality fusion intend to erase the boundaries of different data modalities. Second, the exponential growth of healthcare data provides an opportunity to develop a potent \emph{world model} capable of understanding modalities as special \emph{languages}~\cite{wang2022image}.

In this section, we demonstrate that FM provides a promising avenue to simplify data annotation in healthcare.
For healthcare text annotation, large language models (LLMs) have shown potential capability on clinical language tasks, including medical question answering (Section~\ref{sec:annotation-text}).
For medical imaging, multi-modal FMs and image segmentation FMs improve the efficiency of excessive image captioning and annotation (Section~\ref{sec:annotation-image}).

\subsection{Healthcare Text Annotation}
\label{sec:annotation-text}

Healthcare text annotation is a process that extracts and categorizes critical information from various healthcare texts, playing a crucial role in enhancing data quality and knowledge discovery in the healthcare system~\cite{xia2012clinical}.
However, healthcare text annotation is labor-intensive and time-consuming, as observed from studies in medical condition labeling~\cite{briesacher2008comparison}, patient record annotation~\cite{hogan1997accuracy}, and key medical entity identification~\cite{abacha2011medical}.
To streamline this process, researchers are exploring the use of general-purpose LLMs.
For instance, Gilson et al.~\cite{gilson2022does} reveal that ChatGPT~\cite{chatgpt} achieves human-level performance on medical question-answering tasks, suggesting that LLMs could enhance the efficiency of text annotation more flexibly.
Similarly, Med-PaLM 2~\cite{singhal2023medpalm2}, an LLM fine-tuned with healthcare data, has demonstrated performance comparable to human clinicians in medical knowledge retrieval, reasoning, and question answering.
This indicates that integrating Med-PaLM 2 into medical data annotation workflows could lead to more accurate and efficient annotation processes, harnessing the power of LLMs in the medical domain.

Despite their promise to produce grammatically accurate and human-like text~\cite{thirunavukarasu2023large,gilson2022does}, LLMs still face challenges in specialized healthcare contexts.
A gap in knowledge between general-domain FMs and medical professionals often results in inaccurate responses to realistic patient inquiries, such as in cardiovascular disease prevention~\cite{sarraju2023appropriateness}.
Liao et al.~\cite{liao2023differentiate} further highlight the linguistic differences between medical text generated by ChatGPT and that produced by human experts.

Advancements are required to improve the utilization of text-based FMs for more accurate and efficient medical text annotation.
Recent studies already demonstrate FM-enabled annotation pipelines for pathology instruction synthesis, de-identification, medical graph construction, and terminology-aware dialogue generation~\cite{sun2023pathasst,liu2023deid,wu2024medical,tang2023terminology}. Detailed text-annotation case descriptions are moved to \appref{appendix:ext-annotation}.

\subsection{Medical Image Annotation}
\label{sec:annotation-image}

Medical image annotation includes outlining and labeling anatomical structures, such as organs, tumors, blood vessels, or bones, in histopathology and radiological images~\cite{gao2021utnet,gao2022data}. This process may also involve annotating regions of interest in histopathological slides to identify cancerous cells or specific tissue types of clinical significance.
Medical image annotation is essential in modern healthcare as it helps to extract and interpret valuable information from complex medical image examinations. This is necessary for clinical diagnosis and research, as healthcare professionals require computer-aided algorithms to efficiently identify, search, manage, and interpret disease indicators~\cite{yin2018medical}.

Robust segmentation models can be transformed into powerful tools for healthcare image annotation, reducing annotation costs and facilitating large-scale dataset curation across various healthcare imaging applications~\cite{gao2023training}. For instance, Qu et al.~\cite{qu2023annotating} introduce an AI-driven systematic methodology that accelerates organ segmentation annotation by 500 times.
Beyond segmentation, self-supervised pre-training on large unlabeled histopathology corpora reduces reliance on dense annotations by learning transferable slide/tile representations~\cite{cigao2022contrastivedigital_r20}.
Masked image modeling and retrieval-oriented objectives further support scalable dataset triage and weak supervision for annotation-efficient learning~\cite{filiota2023histopathologymasked_r27,wangx2023retcclclustering_r21}.
Despite advancements in medical image annotation using FMs, the absence of consensus on state-of-the-art benchmarks remains a significant challenge.

The advent of segmentation FMs, especially SAM-like models, has substantially accelerated medical image annotation and interactive segmentation, but domain adaptation remains essential for robust clinical performance~\cite{kirillov2023sam,wang2023sammed,ma2024segment}. We move the detailed SAM-family comparisons and adaptation variants to \appref{appendix:ext-annotation}.

Technical challenges remain in adapting SAM to the medical domain efficiently and scalably.
Medical images, such as X-rays, MRIs, CT scans, and histopathological slides, could reveal domain-specific clinical patterns that require careful interpretation beyond the scope of general-purpose SAM.
In particular, SAM, pre-trained in general vision, lacks the specialized understanding of medical anatomy, pathology, and radiology necessary for precise image analysis.
Data- or parameter-efficient fine-tuning is effective while not scalable. Moving forward, more research efforts are required towards a more scalable medical knowledge transfer pattern for better utilizing and adapting SAM or other general-purpose FMs in medical image and text annotation.

%% file: content_files/7-privacy.tex
\section{Data Privacy}
\label{section:privacy}

Safeguarding healthcare data privacy is crucial given their critical roles in clinical research, diagnosis, treatment, and disease prevention~\cite{gkoulalas2014publishing}. Laws and regulations have been enacted to safeguard these highly sensitive personal records~\cite{clayton2023dobbs, malin2010technical}, which helps to build trust between patients and medical institutions.

Traditional healthcare privacy protection relies on encryption, access control, de-identification, and perturbation mechanisms~\cite{aggarwal2008general,fung2010privacy}. However, linkage attacks and secondary-use risks still challenge practical anonymity in large-scale data sharing~\cite{reiter2019differential}.

Data-centric FMs bring both opportunities and risks: they can synthesize privacy-preserving medical data for downstream development, but large models may also memorize sensitive training information~\cite{hu2023sok,carlini2021extracting}. Recent medical studies show that synthetic and differentially private data generation can improve utility-privacy trade-offs when properly governed~\cite{sei2024local}.

At the same time, memorization-induced leakage remains a major concern for healthcare deployment, especially for protected health information~\cite{carlini2019secret,lehman2021does,sallam2023chatgpt}. Therefore, privacy safeguards are needed across data processing, model training, inference, and deployment.

Promising directions include federated/self-supervised training under institutional governance and risk-aware FM interfaces that proactively detect and suppress privacy-sensitive outputs~\cite{frompretrainingto2025_57bdd335}. Detailed case examples and mitigation workflows are moved to \appref{appendix:ext-privacy}.

%% file: content_files/8-evaluation.tex
\section{Performance Evaluation}
\label{section:evaluation}

Systematic and reliable evaluation is critical for assessing the performance and safety of AI models deployed in healthcare settings.
The evaluation of FMs is challenging owing to their extensive utilization given their model scale and complexity~\cite{chang2023survey-eval}.
In this section, we discuss FM evaluation strategies and challenges from three key aspects, including benchmarking (Section~\ref{sec:evaluation-benchmarking}), human evaluation (Section~\ref{sec:evaluation-human}), and automated evaluation (Section~\ref{sec:evaluation-auto}).

\input{content_files/figures/evaluation_figure.tex}
Figure~\ref{fig:evaluation} overviews the complementary roles of benchmarking, human evaluation, and automated evaluation for healthcare FMs.

\subsection{Benchmarking}
\label{sec:evaluation-benchmarking}

Researchers have established benchmarks for model evaluation in the healthcare field, inspired by the progress of AI research in vision and language domains~\cite{johnson2016mimic-iii,gu2021pubmedbert,wang2023real}.
These benchmarks often build on sample diversity and large quantity scale to facilitate FM evaluation.
For example, the Biomedical Language Understanding Evaluation benchmark (BLUE)~\cite{peng2019bluebert} covers five language tasks with ten biomedical and clinical text datasets, combining PubMed and MIMIC-III~\cite{johnson2016mimic-iii} based applications.
The Biomedical Language Understanding \& Reasoning Benchmark (BLURB)~\cite{gu2021pubmedbert} focuses on biomedical datasets and improves BLUE by including question-answering tasks commonly used to evaluate large language models (LLMs).
Medbench~\cite{liu2024medbench} and recent clinician-curated specialty benchmarks further strengthen domain-specific evaluation for medical LLMs and multimodal systems~\cite{arora2025healthbench,ma2025pathbench}. We move the full benchmark inventory and detailed benchmark-by-benchmark notes to \appref{appendix:ext-evaluation}; representative public datasets are summarized in \appref{appendix:datasets}.
Recent benchmark construction also includes specialty-specific stress testing and multilingual or cross-modality settings, which can expose long-tail clinical failures that may be hidden by aggregate leaderboard scores~\cite{fan2025medodyssey,liu2024expert}.
For dentistry-oriented language understanding, DentalBench adds bilingual tasks and evaluation criteria to cover domain terminology and long-tail dental knowledge for LLM evaluation~\cite{dentalbenchbenchmarkinga2025_27c54b83}.
Complementary clinician-curated and specialty-focused evaluations continue to broaden benchmark coverage for practical clinical settings, including pathology and dermatology tasks~\cite{ma2025pathbench,fan2025medodyssey,huang2025eval,liu2025dermclinical}.

Current benchmarks focus primarily on static datasets, failing to capture the full complexities of how FMs behave in dynamic environments with continuous variants, real-time adjustments, and human interactions.
In addition, FMs are more likely to have seen the common benchmarks during pre-training, potentially leading to biased performance during evaluation~\cite{shi2023detecting}.
From a data-centric perspective, developing dynamic or temporal datasets that reflect complex real-world scenarios could provide a more accurate assessment of FM behavior in healthcare applications.
For \emph{medical agents} that operate over multi-step clinical workflows, evaluation should extend beyond static tasks to sequential clinical simulations, safety and robustness assessments, and system-level metrics that capture workflow impact~\cite{hu2025medicalagentsurvey}.
Interactive environments such as AgentClinic provide step-wise, multimodal simulated clinical workflows for benchmarking agent decision-making over time~\cite{schmidgall2024agentclinic,liu2025interactive}.
For \emph{copilots}, benchmarking should also report clinician-centered outcomes such as decision-time, workload, and calibrated trust alongside task correctness, since these systems are designed to reshape workflows rather than replace them~\cite{mehandru2024agentsclinic,mcduff2025differentialdiagnosis}.

Benchmarks also cover different evaluation metrics for a wide range of tasks.
Conventional evaluation metrics include accuracy, area under the curve (AUC), and mean average precision (mAP) for classification, intersection over union (IoU) for image segmentation, and bilingual evaluation understudy (BLEU) and BERTScore~\cite{zhang2019bertscore} for text-related tasks, which mostly emphasize precise model predictions.
However, FM evaluation largely goes beyond mere accuracy.
Recent work has queried whether benchmark evaluation metrics align with human values~\cite{wornow2023shaky,he2023survey}. The lack of such alignment could pose misleading estimates of FM capabilities, leading to harmful results. In particular, the evaluation of LLM alignment from perspectives of reliability, safety, fairness, explainability, and robustness underscores the importance of FM trustworthiness~\cite{liu2023trustworthy,ding2025aligning}.
Yet, substantial research benchmarks for evaluating FM trustworthiness in healthcare remain lacking.

\subsection{Human Evaluation}
\label{sec:evaluation-human}

In real-world healthcare applications, many scenarios are not covered by major benchmarks (e.g., rare disease assessment~\cite{chen2024rarebench}), so structured expert review remains critical. Representative radiology and differential-diagnosis human evaluation protocols are summarized in \appref{appendix:ext-evaluation}~\cite{chambon2022adapting,mcduff2025differentialdiagnosis}.
In practice, clinician scoring protocols often include report correctness, completeness, and actionability dimensions, and are used to examine failure modes not reflected by automatic metrics alone~\cite{lyu2023gpt-translate-radiology,peng2023study}.
For example, Chambon et al.~\cite{chambon2022adapting} assess the clinical correctness of FM-generated chest X-ray images with the help of radiologists.
Lyu et al.~\cite{lyu2023gpt-translate-radiology} invite two experienced radiologists to evaluate the overall quality, completeness, and correctness of LLM-translated radiology reports.
Peng et al.~\cite{peng2023study} invite two physicians to perform a Turing evaluation on 30 paragraphs written by FMs and humans, respectively, to assess readability and clinical relevance.
Moor et al.~\cite{moor2023med-flamingo} implement a human evaluation application for clinical experts to rate the LLM-generated answers for medical image-related questions.
Recent clinician studies of LLM-assisted differential diagnosis further compare unassisted and assistive performance on challenging real-world cases sourced from published case reports~\cite{mcduff2025differentialdiagnosis}.

Introducing human evaluation to data curation and model training can align model outcomes more closely with human values.
For instance, Reinforcement Learning from Human Feedback (RLHF)~\cite{ouyang2022training} combines human evaluation with reinforcement learning techniques to train language models that exhibit strong alignment with user instructions, which serves as the cornerstone of modern chatbots like ChatGPT~\cite{chatgpt}.
This approach also adopts human ranking to evaluate the final model performance.
Similarly, in the healthcare field, IvyGPT~\cite{wang2023ivygpt} provides rich diagnosis and treatment answers by applying the RLHF technique on medical question-answering data.

However, relying on human experts' assessment can incur high costs and introduce subjectivity and bias, particularly given the flexibility of FM outputs.
Training and hiring healthcare experts can be expensive~\cite{van2022critical}.
Moreover, variations in experts' backgrounds, experiences, motivations, and methods of inquiry make it difficult to achieve uniform assessments~\cite{gehrmann2023repairing}.
Efforts have been made to address these challenges. For example, Tam et al.~\cite{tam2024framework} propose QUEST, a comprehensive and practical framework for human evaluation of LLMs in healthcare, with a particular emphasis on the significance of blind assessment in ensuring robust and unbiased evaluations.

\subsection{Automated Evaluation}
\label{sec:evaluation-auto}

Leveraging strong and well-aligned FMs for automated evaluation is promising, including reference-free scoring and fine-grained skill-level judging~\cite{chen2023exploring,chiang2023can,jain2023sse,ye2023flask}. In healthcare, this line remains underexplored, and detailed examples are deferred to \appref{appendix:ext-evaluation}, including case-based setups that complement automated clinical reasoning assessment~\cite{mcduff2025differentialdiagnosis,tu2025conversationaldiagnostic}.
For instance, Chen et al.~\cite{chen2023exploring} explore reference-free evaluation methods that prompt ChatGPT to score the quality of model-generated texts without a pre-defined ground truth.
Chiang et al.~\cite{chiang2023can} show that LLM-based evaluation remains stable across prompt formats and aligns well with human expert assessments.
Jain et al.~\cite{jain2023sse} propose self-supervised evaluation strategies to assess LLM properties without benchmarks or human annotations.
Ye et al. introduce FLASK~\cite{ye2023flask}, a fine-grained LLM evaluation protocol that decomposes coarse-level scoring into instance-wise skill assessments. FLASK prompts GPT-4 to assign scores with rationales for specific alignment skills of target FM-generated text.
From a data-centric perspective, such benchmark FMs with extensive pre-training knowledge and human alignment can be used as supplementary evaluators for other FMs.
Related studies on differential diagnosis and conversational diagnostic AI provide structured case-based evaluation setups that can complement automated evaluation of clinical reasoning in realistic healthcare settings~\cite{mcduff2025differentialdiagnosis,tu2025conversationaldiagnostic}.
While FMs have demonstrated the ability to understand and generate medical text or images~\cite{nori2023gpt4-on-medical,roy2023sam-md,yu2023mm-vet}, they exhibit limitations including inconsistent performance across datasets \cite{cheng2023sam} and lower accuracy compared with state-of-the-art in-domain learning methods \cite{he2023accuracy}.
Whether automated evaluation via FMs is trustworthy for healthcare applications remains an unresolved question.
An integrated approach that combines automated evaluation with human expert supervision could potentially yield more reliable results.
Moving forward, the development of rigorous, healthcare domain-specific evaluation benchmarks for imaging, text, and genomics is essential. These benchmarks should be thoughtfully designed, deployed, and validated to ensure their effectiveness in healthcare contexts.

%% file: content_files/figures/evaluation_figure.tex
\begin{figure}[t]
    \centering
    \includegraphics[width=\linewidth]{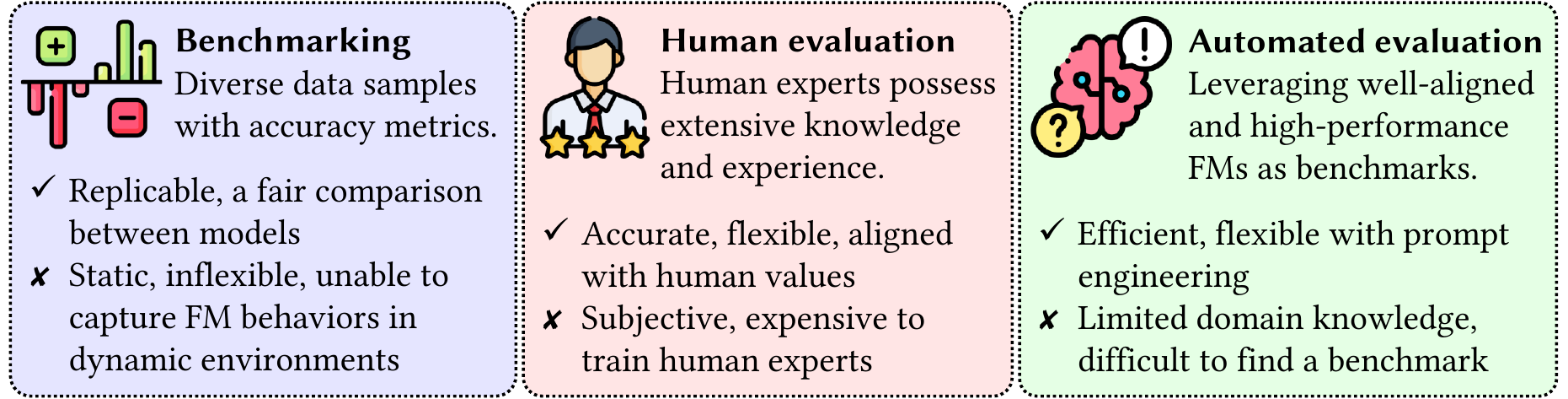}
    \caption{Foundation model evaluation strategies.}
    \label{fig:evaluation}
\end{figure}

%% file: content_files/9-outlook.tex
\section{Challenges and Opportunities}
\label{section:limitation}

Despite the promise of FMs in healthcare, open challenges remain in developing and adapting trustworthy FMs to gain insights into patient outcomes and clinical workflow. We discuss key directions for building reliable healthcare-focused FMs towards better human-AI alignments, addressing hallucination, bias, and regulation.

\subsection{Healthcare-Focused Foundation Model Development}

General-purpose FMs still face substantial domain gaps when transferred to clinical data and workflows~\cite{yang2026scalar}. In practice, fine-tuning and adapter-based adaptation are often necessary, but they must balance efficiency against catastrophic forgetting and unstable cross-task generalization~\cite{zhang2023challenges}. Robust benchmark design for tuning strategies remains a prerequisite for reliable healthcare deployment.

Healthcare FM development also requires stronger multimodal integration across imaging, text, and biological data to capture multi-scale disease patterns and improve patient-level decision support~\cite{gao2023training}. Detailed architecture- and modality-level discussion is moved to \appref{appendix:ext-outlook}.

\subsection{Hallucination}

Hallucination refers to generating plausible but inaccurate medical content, which can directly affect diagnosis, triage, and treatment decisions~\cite{ref7}. Promising mitigation combines verifiable reasoning-oriented post-training, retrieval grounding (including medical-graph retrieval), and explicit uncertainty communication~\cite{guod2025deepseekr1_r05,lewis2020retrieval}.

In addition, hallucination control should be validated with medical-domain benchmarks, human review, and adversarial stress testing before real-world deployment~\cite{ref8}. Detailed benchmark structures and case-level analysis are provided in \appref{appendix:ext-outlook}.

\subsection{Bias}

Bias in healthcare FMs can originate from skewed data distributions, model inductive biases, and misaligned post-training objectives~\cite{ferrara2023should}. Existing studies already report language- and population-dependent performance disparities in clinical evaluations, indicating that fairness remains unresolved under data-efficient transfer~\cite{sun2025fairness,queiroz2024does}.

Addressing bias requires geographically and demographically broader data curation, stronger dataset auditing, and systematic human-in-the-loop evaluation protocols~\cite{tam2024framework}. More detailed examples and evaluation notes are moved to \appref{appendix:ext-outlook}.

\subsection{Regulation}

Healthcare FM deployment requires governance that aligns model capability with legal and clinical accountability (e.g., HIPAA/FDA-related requirements)~\cite{ref4}. Beyond model-level metrics, workflow-integrated medical-agent systems need auditability, safe fallback, and continuous monitoring of system-level outcomes under distribution shift~\cite{hu2025medicalagentsurvey}.

Open-source transparency and data standards remain critical for trustworthy regulation and reproducibility, including metadata templates and interoperability principles such as DICOM/PACS, CEDAR, FAIR, and dataset datasheets~\cite{tan2025medforge,ref11}. A detailed regulation checklist and operational recommendations are provided in \appref{appendix:ext-outlook}.

%% file: content_files/10-conclusion.tex
\section{Conclusions}
\label{section:conclusion}

The striking progress of foundation models (FMs) and their applications in healthcare open up possibilities for better patient management and efficient clinical workflow. In these efforts, collecting, processing, and analyzing scalable medical data has become increasingly crucial for FM research. In this survey, we have offered an overview of FM challenges from a data-centric perspective. FMs possess great potential to mitigate data challenges in healthcare, including data imbalance and bias, data scarcity, and high annotation costs. Due to FM's strong content generation capabilities, there is a remarkable need for greater vigilance regarding data privacy, data bias, and ethical considerations about the generated medical knowledge. Only by adequately and reliably addressing the data-centric challenges can we better leverage the power of FMs across a broader scope of medicine and healthcare.

%% file: content_files/appendix.tex
\section{Healthcare Data Modalities}
\label{appendix:modality}

There are a plethora of healthcare data modalities collected in different clinical settings, such as imaging, clinical notes, biosensor records, and electroencephalography. These data records provide multi-scale information for professionals in clinical diagnosis, prognosis, and treatment development~\cite{azam2022review,kumar2020whole,heikenfeld2018wearable}.
Patient diagnosis is routinely based on analyzing complex disease patterns derived from various healthcare data modalities. The diversity of medical modalities comes from the distinct methods of data acquisition, including the use of invasive procedures or the types of medical devices. We discuss common healthcare data modalities in clinical practices, including radiology, histopathology, molecular information, clinical information, blood testing, biosensors, and electroencephalography.

\emph{Radiology} commonly includes X-ray, magnetic resonance imaging (MRI), computed tomography (CT), positron emission tomography (PET), and ultrasound scans. From radiological image inputs, we are able to capture visual patterns that are helpful for clinicians to make diagnoses, providing valuable information about the anatomical structure and function of the body's internal organs and tissues~\cite{azam2022review,wang2012machine}. Various imaging principles, such as ionizing radiation, radio waves, gamma rays, and sound waves, are utilized to visualize the internal structures and activities non-invasively~\cite{herring2019learning}.
These radiographs provide key information about the body's tissues, organs, and bones as well as metabolic processes that are crucial to assess disease characteristics.
Radiological images play an important role in the application of FMs in healthcare~\cite{sowrirajan2021moco-cxr,tiu2022expert}. In particular, radiological images and reports can be analyzed in a joint-modal manner, providing a more comprehensive understanding of patient conditions~\cite{chambon2022adapting,zhang2022contrastive}.

\emph{Histopathology} involves examining visually perceptible changes in cells and tissues to study the manifestations of diseases~\cite{kumar2020whole,kumar2019multi}. In clinical practice, tissue is first removed from the body, and then the histological sections are placed onto glass slides. A high-resolution slide scanner equipped with microscope optics serially captures image tiles across the tissue section. The seamless reconstruction of these tile images yields a single monolithic digital representation of the entire glass slide, typically within the order of billions of pixels.
Staining techniques highlight architectural and morphological details, demonstrating the changes in cells and the actual causes of the illness. Therefore, the microscopic examination of stained tissue sections offers a definitive disease diagnosis.
Different magnifications in histopathology imaging can reveal morphological details, offering multi-level visual perceptions in clinical scenarios. Higher magnifications reveal finer attributes, while lower magnifications facilitate the understanding of overall tissue morphology~\cite{ashtaiwi2022optimal}.
Due to the high resolution of histopathological images, current FMs mostly take split image patches as input instead of whole slides~\cite{huang2023plip,lu2024visual}.

\emph{Molecular information} typically encompasses various categories of biomarkers, including DNA, RNA, proteins, metabolites, and, more recently, microbiome profiles~\cite{mishra2010cancer}. These molecular signatures have remarkably aided disease diagnosis, prognosis, treatment, and monitoring of therapeutic response across cancer, infectious disease, and genetic disorders.
Molecular diagnostics analyzes biofluids or tissue samples to identify abnormalities at the genetic and molecular level, which allow disease signature detection and target therapy planning~\cite{subramanian2020multimodal,lu2005microrna}.
However, interpreting complex multidimensional molecular data requires specialized bioinformatics pipelines, particularly from the perspective of FM~\cite{cui2023scgpt}. While molecular profiles offer valuable insights into disease formation and progression, data-driven analytics and clinical trial validation are increasingly necessary before the widespread deployment of molecular testing into real-world healthcare utility.

\emph{Clinical information} typically encompasses structured metadata documenting patients' demographic details, medical history, interventions, outcomes, and other relevant variables~\cite{simpao2014review}. Sources of clinical data include electronic health records (EHR), clinical trial case report forms, disease registries, and health surveys. These structured fields capture details such as age, gender, medications, procedures, family history, social factors, and clinician notes. Such metadata complements diagnostic results to provide crucial contexts for enhancing patient care. Clinical information can aid patient stratification, disease pattern recognition, healthcare predictive analytics, and decision-making optimization in precision medicine applications~\cite{shaikhina2019decision}.
Meanwhile, clinical text and imaging records stand as the frontier data modalities for FM development. This underscores the importance of clinical information in the advancement of healthcare FMs~\cite{alsentzer2019clinical-bert}.
However, heterogeneity across institutions and populations poses data integration challenges, such as the documentation conventions of various institutions as well as individual writing styles. Standardization of clinical terminologies and ontologies is thus essential for unambiguous data sharing and mining~\cite{kim2019developing}. As a result, multi-scale clinical data records could complement other data modalities to enable a holistic understanding of patients and personalized therapies.

\emph{Blood testing} reveals rich information by analyzing constituents of blood specimens like cell counts, proteins, metabolites, and hormones in patient blood samples~\cite{burtis2014tietz}. The specimens are usually obtained through minimally invasive venipuncture or fingerprick and are analyzed using automated systems. Analysis techniques encompass microscopy, flow cytometry, chromatography, etc~\cite{Li2022bloodtest}. The analysis of blood testing for patients is usually based on the assessment of biomarker levels, which reflect patient health status and assist in screening, diagnosis, treatment, and monitoring across a wide range of conditions~\cite{brusselle2018blood,hier2021blood}. AI techniques have been explored for the discovery of blood biomarkers in cancer~\cite{visaggi2021modern}.
Another example is MediTab~\cite{wang2023meditab}, an FM for medical tabular data prediction that scales up model pre-training with blood testing data.
However, FM for tabular data analytics, including blood testing data, remains challenging due to the difference between the structured data and free-form text~\cite{bisercic2023interpretable}.

\emph{Biosensors} are devices or probes containing a biological recognition element coupled to a transducer that converts a physiological or biochemical signal into a measurable electronic output~\cite{haleem2021biosensors}. Biosensing techniques typically include electrochemical, optical, piezoelectric, thermal, and magnetic detection~\cite{naresh2021review}. Biosensors allow continuous tracking of vital signs, physiological signals, and biomarkers by transducing biological responses into electrical outputs~\cite{heikenfeld2018wearable}.
These multi-modal data streams from devices monitoring heart rates and glucose levels could enable personalized medicine through remote patient surveillance and assessment of health status changes~\cite{majumder2017wearable}. In particular, biosensors offer great promise for ubiquitous real-time health monitoring, thus providing extra information in multi-modal computer-aided diagnosis.

\emph{Electroencephalography} (EEG) measures electrical brain activity by recording voltage fluctuations resulting from ionic currents within neurons~\cite{steriade2005cellular}. Electrodes are placed on the scalp to record electrical activity emerging from firing neurons in the brain that manifest as brain waves with distinct rhythmic patterns~\cite{van2015opportunities}. Analysis of brain waves can facilitate the diagnosis of neurological disorders, concussions, sleep abnormalities, and other conditions~\cite{craik2019deep}. Wearable EEG devices can facilitate longitudinal brain monitoring to assess disease progression, recovery, and treatment effects.
Research has demonstrated that self-supervised EEG representation learning with massive unlabeled EEG signals for few-shot sleep staging tasks is feasible~\cite{yang2023self}.
However, signal artifacts, low spatial resolution, and inter-subject variability are ongoing research challenges. Overall, EEG signals offer a non-invasive access to correlate brain physiology with function, behavior, and pathology, which greatly facilitates clinical diagnosis and healthcare research.

\section{Extended Case Compendium}
\label{appendix:extended-case-compendium}

This section consolidates detailed domain-wise examples moved from the main text to preserve citation coverage while keeping the core narrative concise.

\subsection{Extended Multi-Modal Case Studies}
\label{appendix:ext-modality}

\paragraph{Joint-modal pre-training case details.}
Beyond early vision-language pre-training (VLP) in specific imaging domains, joint-modal and self-supervised pre-training have begun to yield clinical-grade vision foundations across diverse imaging specialties, spanning both clinically deployed systems and open-source releases with reusable benchmark suites, and ranging from single-modality scaling to cross-modal alignment with reports and clinical knowledge. From a data-centric perspective, these advances are increasingly enabled by curated real-world corpora, harmonized preprocessing, and multi-site evaluation that quantify distribution shifts and failure modes.

In pathology, patch- and slide-level self-supervised pre-training on real-world data has enabled generalist representations spanning patch-level foundations and hierarchical tile-to-slide learning, while clinical-grade and general-purpose pathology FMs are moving toward broader clinical coverage and task diversity~\cite{chen2023uni,xu2024whole,chenrj2024computationalpathology_r07,vorontsove2024clinicalgrade_r08,zhou2026ai,lei2025unifying,zheng2025graphmamba}.
Pathology FMs are also moving toward unified multimodal patch/WSI modeling, stain-aware analysis, and slide-report alignment to support diverse slide conditions~\cite{suny2025cpathomni_r18,huas2024pathoduetpathological_r23,ding2025titan}.
For precision oncology, MUSK aligns histology with clinical context for decision support~\cite{xiangj2025languageprecision_r15}.
Cross-site generalization remains an active focus, motivating robust pre-training and distillation frameworks~\cite{maj2025generalizablepathology_r11}.
At scale, broad task suites and diverse downstream coverage highlight the move toward more general pathology foundations~\cite{yanf2025pathorchestracomprehensive_r12}.
Open-source pathology visual foundations and evaluation suites are also emerging, lowering barriers for community study and reproducibility, and making it easier to probe clinically important long-tail failure modes~\cite{vorontsove2024clinicalgrade_r08,modelscopendruipath_r37,ma2025pathbench}.
From a data-centric view, pathology FM training is increasingly enabled by larger and more heterogeneous corpora and stronger backbones, while downstream evaluation is becoming more diverse across tasks and clinical settings. Representative directions include patch-level generalist pre-training, mixed-magnification self-supervised scaling, and hierarchical tile-to-slide learning, which together better model the multi-scale heterogeneity inherent to WSI data~\cite{chen2023uni,zimmermanne2024virchow2mixed_r36,xu2024whole}.

Beyond fundus images, multimodal and knowledge-enhanced pre-training supports OCT analysis and report-grounded retinal representation learning~\cite{multimodalfoundationmode2025_18940613,retclipa2024_2dfb24e5,mmretinalknowledge2024_39021b7e}.
Clinical knowledge augmentation and multimodal generalist modeling further broaden ophthalmic imaging coverage~\cite{visionuniteavision2025_1d6720a2,eyefoundamultimodal2024_3b6dad82}.
Integration of imaging with downstream care pathways has shown promise for diabetes screening and primary care decision support~\cite{integratedimagebased2024_10a30804}.
Computational ophthalmology is also exploring vision-language foundations beyond fundus images~\cite{amultimodalvisual2025_a4db3fca}.
In ophthalmology, multimodal multitask vision FMs support generalist ophthalmic AI and clinical assistance across populations~\cite{developmentandvalidation2024_0437bee7,aneyecarefoundation2025_e5f2b10e}.

Dermatology models similarly integrate multimodal supervision and large-scale training for clinical decision support~\cite{amultimodalvision2025_803e276d}.
Dermatology research also explores multimodal assistants and knowledge-enhanced pre-training for assessment and differential diagnosis~\cite{pretrainedmultimodal2024_e7906fe2,makemultiaspect2025_af7346b7,derminohybridpretraining2025_a9a4ceca}.

Recent radiology foundation models explore open-world segmentation~\cite{ageneralistfoundation2025_472e98b5} and 3D vision-language modeling for robust generalization~\cite{visionlanguagefoundation2025_94a6ee60}.
For volumetric imaging, self-supervised pre-training and vision-language foundations have been developed for CT and MRI~\cite{ageneralizable3d2025_57d7aeb9,merlinavision2024_2f35dbf0,triadvisionfoundation2025_28d8cd7f}.
Multimodal generative models enable flexible interpretation of 3D images and videos~\cite{multimodalgenerativeai2025_e450cd00}, while community perspectives emphasize open evaluation and reproducibility~\cite{pillar0a2025_dc4bb63c}.
In chest imaging, large-scale self-supervision supports general CXR understanding~\cite{evaxa2025_5859b0b4}, and vision-language modeling enables CT report generation and outcome prediction~\cite{visionlanguagemodel2025_437b71a9}.
Domain-specific CT foundation models are being developed for lung and colorectal cancer imaging~\cite{alungct2025_e72afd13,crcfoundacolorectal2025_0789b6be,lei2025data}.
For screening modalities such as mammography, specialized foundation models target robust interpretation at scale~\cite{aversatilefoundation2025_080f3c4a}.

Ultrasound studies also move toward scalable video modeling and report-grounded representation learning for echocardiography~\cite{christensen2024echoclip,echovisionfm2025_273c1b70}.
Multi-video view modeling further improves comprehensive echocardiography interpretation~\cite{echoprimeamulti2024_04fed096}, while generalizable echo foundations support downstream analysis and deployment~\cite{echofmfoundationmodel2025_45c40046}.
Open and generalizable ultrasound foundation models facilitate broader community evaluation and deployment~\cite{openusafully2025_e96804ca,afullyopen2025_448ab695}.

In dentistry, multimodal vision-language assistants and cross-modality contrastive pretraining across CBCT and intraoral scans support comprehensive diagnosis and tooth segmentation~\cite{dentvlmamultimodal2025_a3531ade,multimodalcontrastivepre2025_7ecc89b6}.

\paragraph{LLM-driven fusion and modality-expansion case details.}
In domain-specific settings, multimodal copilots and multimodal LLM agents are being explored to provide interpretable, interactive assistance, such as pathology diagnostic reasoning agents~\cite{lu2024multimodal,suny2025cpathagentagent_r19,chenc2025evidencediagnostic_r17}.
Conversational assistants have also been explored for patient-facing ophthalmology support~\cite{retinalgptaretinal2025_362d6b76}.
In dentistry, multimodal LLMs target oral-health reasoning and conversational workflows~\cite{dentalgptincentivizingmu2025_7f9c9e55,oralgptomnia2025_13c2468e}.

Current multi-modal FMs predominantly focus on language and vision, given their ubiquitous use as the most common modalities.
A unifying theme is that these non-vision modalities are often temporally structured and weakly labeled, so data curation, alignment, and evaluation become as critical as model design for safe deployment.
Yet, emerging work expands FMs to biosignals, audio, and video (e.g., ECG~\cite{ecglmunderstanding2025_a9510e52}, respiratory audio~\cite{respllmunifyingaudio2024_5058e290}, and endoscopy video~\cite{foundationmodelfor2024_c3f626c8}).
Large-scale self-supervised video FMs are also being explored for operating-room understanding and intelligent surgery~\cite{largescaleself2025_8264afeb}.
Specialized ECG foundations can provide robust waveform representations and unified diagnosis-and-explanation capabilities~\cite{practicalintelligentdiag2023_31168f43,foundationmodelof2024_ae2e3385}.
LLM-integrated systems further improve interpretability and interaction for ECG diagnosis~\cite{ecgdoctoran2025_69d8dd57}.
Beyond ECG, biosignal foundation models extend to digital stethoscope signals and multimodal cardiac sensing at the population scale~\cite{foundationmodelsfor2024_f30d2a1c,sensingcardiachealth2025_f4f62783}.
Specialized 3D cardiac imaging foundation models are also emerging for CT-based clinical workflows~\cite{acardiacspecific2025_060dac1a}.
Guideline-centric reasoning agents are also explored for echocardiography measurement and interpretation~\cite{echoagentguidelinecentri2025_50a64f23}.
In neuroscience, foundation models have been explored for brain activity recordings~\cite{caro2023brainlm} and neuroimaging applications spanning 3D brain MRI and radiology report generation~\cite{towardsgeneralisablefoun2025_a3029089,towardsaholistic2025_51ae3d1a,rui2024brainmvp,wang2024brainsck}.
Recent work further explores unified EEG/MEG foundations and generalist neuroimaging learning at health-system scale~\cite{brainomniabrain2025_3bd515fa,healthsystemlearning2025_d5b7b655}.
Head CT disease detection also benefits from emerging 3D foundation models tailored to radiology workflows~\cite{ref3dfoundationai2025_69571ed6}.
For FM applications in the healthcare field, there are also examples including molecules~\cite{liu2023moleculestm}, genomics~\cite{ding2023pathology}, and tabular data~\cite{wang2023meditab}.

\subsection{Extended Adaptation and Pre-training Examples}
\label{appendix:ext-medfm}

Gema et al.~\cite{gema2023peft-llama} evaluate the language modeling performance of various PEFT methods, demonstrating that LoRA, among the major PEFT approaches, works most effectively for clinical domain adaptation. Zu et al.~\cite{zu2024embedded} introduce a perspective recognizing prompt as a distribution calibrator and propose Embedded Prompt Tuning, an enhanced prompt tuning method, which outperforms other PEFT methods in medical few-shot classification tasks. They demonstrate that prompt tuning mitigates the domain shift between FM pre-training data and medical data. Besides, Dutt et al.~\cite{dutt2023fairtune} highlight that the optimal PEFT strategy depends on data and the generalization objective. They propose the Fairtune framework to optimize the choice of PEFT strategies. Notably, systematic comparisons of PEFT methods in healthcare remain largely limited; insights gained from general domain research can be considered. For instance, Ding et al.~\cite{ding2023parameter} show that adapters facilitate multi-task fine-tuning, prompt-based methods excel in low-data scenarios despite optimization challenges, and LoRA achieves fine-tuning-level performance without inference latency.

Contrastive learning is another important pre-training strategy for medical FMs. For example, REMEDIS~\cite{azizi2023robust} is a medical vision model pre-trained via contrastive learning to extract representative visual features for medical images.
Pai et al. develop an FM for cancer imaging biomarker discovery by contrastively training a convolutional vision encoder~\cite{pai2024foundation}.
Vision-language models such as MI-Zero~\cite{lu2023mi-zero}, PLIP~\cite{huang2023plip}, CONCH~\cite{lu2024visual}, and MONET~\cite{kim2024transparent}, are contrastively pre-trained on domain-specific image-text pairs. They achieve positive performance on zero-shot image classification tasks in radiology, pathology, and dermatology.

For example, BioBERT~\cite{lee2020biobert} is pre-trained on PubMed abstracts and PubMed Central (PMC\footnote{\url{https://www.ncbi.nlm.nih.gov/pmc/}}) full-text articles with BERT initialization and demonstrates better performance compared with BERT and previous state-of-the-art models when fine-tuned on three biomedical language tasks.
PMC-LLaMA~\cite{wu2023pmc-llama} is an open-source language model fine-tuned from LLaMA~\cite{touvron2023llama} on biomedical academic papers.
Singhal et al. ~\cite{singhal2022med-palm} apply instruction tuning on Flan-PaLM~\cite{chung2022flan-palm} to obtain the Med-PaLM model, achieving state-of-the-art performance on a broad range of medical question answering (MedQA) tasks.
MMedLM~\cite{qiu2024mmedc} is a multi-lingual language model for medicine, further pre-trained from InternLM~\cite{cai2024internlm2} using over 25B medical-related tokens across six languages.
PubMedCLIP~\cite{eslami2021pubmedclip}, a CLIP model fine-tuned on medical image-text pairs from PubMed articles, shows outstanding performance on medical visual question answering (MedVQA) tasks.

\subsection{Extended FM Fundamentals and Prompting Details}
\label{appendix:ext-fm-basics}

\paragraph{Scaling examples across language and vision.}
Classic language scaling examples include GPT~\cite{radford2018gpt}, BERT~\cite{devlin2018bert}, T5~\cite{raffel2020t5}, GPT-3~\cite{brown2020gpt3}, and PaLM~\cite{chowdhery2022palm}, which progressively increase pre-training corpus size and model parameters. Similar trends are observed in vision and multimodal FM development, such as DALL$\cdot$E~\cite{ramesh2021dalle}, CLIP~\cite{radford2021clip}, SAM~\cite{kirillov2023sam}, and scalable visual pre-training objectives like BEiT~\cite{wangw2023imageforeign_r30}.

\paragraph{Self-supervised objective details.}
Input-reconstruction objectives include masked language modeling (e.g., BERT~\cite{devlin2018bert}), masked image modeling~\cite{he2022mae,xie2022simmim}, and autoregressive next-token prediction (e.g., GPT~\cite{radford2018gpt}). Contrastive objectives include SimCLR and MoCo-v3~\cite{chen2020simple,chen2021mocov3}, as well as image-text alignment in CLIP~\cite{radford2021clip}. Hybrid objectives continue to evolve, such as DINO-style pre-training~\cite{oquab2023dinov2,simeoni2025dinov3}. In healthcare, SSL pre-training supports data-efficient biomedical adaptation, including PubMedBERT and MoCo-CXR~\cite{gu2021pubmedbert,sowrirajan2021moco-cxr,krishnan2022self}.

\paragraph{PEFT and prompting strategy examples.}
Representative PEFT algorithms include BitFit, adapter tuning, prompt/prefix tuning, and LoRA~\cite{zaken2021bitfit,houlsby2019adapter,lester2021prompt-tuning,li2021prefix-tuning,hu2021lora}. Representative ICL strategies include chain-of-thought, self-consistency, tree-of-thought, and self-refinement~\cite{kojima2022cot,wang2022cot-sc,yao2023tot,madaan2023self-refine}. In medical applications, prompting and retrieval are often combined for translation, segmentation guidance, and evidence-grounded reasoning~\cite{lyu2023gpt-translate-radiology,roy2023sam-md,zhu2025guiding,zhu2024realm,wu2024medical}.

\subsection{Extended Data Quantity Cases}
\label{appendix:ext-quantity}

In medical image data augmentation, diffusion models have shown flexibility and scalability, enabling the generation of synthetic data to supplement healthcare datasets~\cite{kazerouni2023diffusion}.
This approach enhances information by drawing from the broader domain of general visual knowledge.
The effectiveness of diffusion-based data augmentation is notably evident across various imaging modalities, including chest X-rays (CXRs)~\cite{chambon2022adapting}, computed tomography (CT)~\cite{liu2022dolce}, brain magnetic resonance imaging (MRI)~\cite{pinaya2022brain}, histopathology~\cite{moghadam2023morphology}, and dermatology~\cite{akrout2023diffusionbased}.
Despite the inherent modality-specific variations, FMs can transfer the knowledge derived from general vision to diverse healthcare contexts.
Beyond diffusion, vision-language foundation models have also been explored for generating realistic radiographs as controllable synthetic data for chest imaging~\cite{avisionlanguage2024_502e1fd2}.
For instance, Pinaya et al.~\cite{pinaya2022brain} harness latent diffusion models~\cite{rombach2022high} from general vision to generate synthetic images from high-resolution 3D brain scans.
Furthermore, Sagers et al.~\cite{sagers2022improving} showcase the potential of DALL$\cdot$E 2~\cite{ramesh2022hierarchical}, a text-to-image diffusion model, to generate realistic depictions of skin diseases across varying skin types.

Meanwhile, FMs demonstrate positive performance in clinical text mining through textual data augmentation. Employing ChatGPT~\cite{chatgpt} to generate high-quality synthetic data with labels has proved beneficial for fine-tuning local models for downstream tasks~\cite{tang2023does}. This approach effectively addresses the challenge of data quantity in clinical text mining. Similarly, PathGen-1.6M provides a large-scale synthetic pathology image--text corpus that can support vision-language pre-training~\cite{sun2024pathgen16m}.
Similarly, Yuan et al.~\cite{yuan2024large} leverage ChatGPT to modify lexical choice and syntax while maintaining semantic meaning, thereby enhancing patient-trial matching data for downstream fine-tuning.
Using ChatGPT to generate synthetic data significantly reduces the burden of extensive data collection.
Recent studies also explore advanced synthesis techniques such as Mamba-enhanced UNets and cycle-consistent diffusion for highly realistic CT and MRI generation~\cite{song2024cycle,zhou2025glfc,song2024cross}.

In addition to general FMs, healthcare-focused FMs offer a reasonable direction for pursuing data efficiency in downstream healthcare applications.
To illustrate, REMEDIS~\cite{azizi2023robust} combines large-scale supervised transfer learning with self-supervised learning to reduce data requirements.
It achieves comparable performance to fully supervised models in out-of-distribution settings using only 1-33\% of downstream data.
Similarly, Dippel et al.~\cite{dippel2024ai} train an unsupervised anomaly detection FM that outperforms supervised models, especially for rare diseases.
BrainSegFounder~\cite{cox2024brainsegfounder}, a neuroimage segmentation FM pre-trained with generally healthy brain MRI images, shows great potential in reducing the data required for downstream fine-tuning.
As for medical text data, Mishra et al.~\cite{mishra2023improving} demonstrate that pre-trained medical text encoders excel at handling low-prevalence diseases, highlighting their potential to address data quantity challenges across healthcare domains.
Moreover, Yi et al.~\cite{yi2023general} introduce continual learning, including sequential learning and rehearsal learning, based on medical FMs as a practical, data-efficient learning paradigm to tackle data quantity challenges effectively.

One common approach is to extract text and images from PubMed Central (PMC), an Internet-scale archive of biomedical literature, or other well-established medical databases.
For instance, PMC-OA~\cite{lin2023pmc-clip} and PMC-15M~\cite{zhang2023large} extract image-caption pairs from PMC articles.
Virchow~\cite{vorontsove2024clinicalgrade_r08} collects pathology scans from the Memorial Sloan Kettering Cancer Center (MSK), illustrating institution-scale data curation from a clinical research partner.
Other examples leverage large online platforms. For example,
OpenPath~\cite{huang2023plip} utilizes pathology hashtags to collect pathology-related tweets with certain keywords or high numbers of likes.
Quilt-1M~\cite{ikezogwo2023quilt-1m} searches YouTube for histopathology videos and obtains corresponding text through speech-to-text techniques and LLM postprocessing.
HealthCareMagic-100k~\cite{yunxiang2023chatdoctor} collects patient-physician dialogues from an online medical consultation platform.

\subsection{Extended Annotation Cases}
\label{appendix:ext-annotation}

PathAsst~\cite{sun2023pathasst} leverages FMs as a generative AI assistant to transform predictive analytics in pathology.
Using ChatGPT and GPT-4, PathAsst generates over 180,000 instruction-following samples to invoke pathology-specific models and facilitate effective interactions based on user inputs and images.
Also, Liu et al.~\cite{liu2023deid} develop DeID-GPT, a GPT-4-based framework for automatically de-identifying clinical text records.
Wu et al.~\cite{wu2024medical} construct a medical knowledge graph by prompting an LLM to identify and extract entities from segmented medical documents. The LLM is further employed to detect relevance and link the entities, forming a three-level hierarchy.
Moreover, Tang et al.~\cite{tang2023terminology} utilize a language model to enhance medical dialogue generation by focusing on domain-specific terminology, demonstrating improved efficiency in generating responses based on the dialogue between medical professionals and patients.

The advent of segmentation FMs, such as the Segment Anything Model (SAM)~\cite{kirillov2023sam}, has significantly advanced automated image segmentation and annotation. Trained with over 1 billion masks sourced from 11 million natural images, SAM can perform zero-shot image segmentation using diverse input prompts, including masks, boxes, and points. Leveraging SAM for medical annotation presents a promising direction for analyzing complex medical objects.
Numerous studies have explored FM-based medical image segmentation with notable results.
Wang et al.~\cite{wang2023sammed} present $\mathrm{SAM^{Med}}$, an enriched framework for incorporating vision foundational models into medical image annotation.
Hu et al.~\cite{hu2023sam-tumor} propose a SAM-based approach for multi-phase liver tumor segmentation, highlighting its efficacy as a robust and interactive annotation tool.
He et al.~\cite{he2023accuracy} summarize SAM's performance across 12 public medical image segmentation datasets, encompassing diverse organs, imaging modalities, and health conditions.
Their findings indicate that pre-trained SAM, without fine-tuning on medical data, underperforms deep learning models specifically trained on medical tasks. Additionally, Mazurowski et al.~\cite{mazurowski2023segment} evaluate SAM on 19 distinct medical imaging datasets, revealing its promising zero-shot medical segmentation capabilities.

Although SAM is a promising segmentation-focused FM, the inherent domain gap between general vision and medical imaging remains to be addressed.
Huang et al.~\cite{huang2023segment} explore various evaluation strategies for SAM, analyzing the factors that influence its segmentation performance. Their insights are essential for refining SAM's application in the medical context. However, the consistency and reliability of directly applying SAM in medical image segmentation still require improvement through careful refinement of FMs~\cite{zhang2023segment}.
Several studies have focused on adapting FMs to the complexities of medical data to improve segmentation performance, which potentially benefits image annotation.
Zhang et al.~\cite{zhang2023input} propose SAMAug, which incorporates input augmentation by integrating prior maps with raw images to enhance medical image segmentation.
MedSAM exemplifies the process of fine-tuning SAM for medical images~\cite{ma2024segment}, with the ultimate goal of developing a versatile tool applicable to a wide range of medical tasks.
Similarly, Med SAM Adapter~\cite{wu2023medical} introduces a practice approach by infusing medical-specific domain knowledge into SAM, significantly enhancing its performance in medical image annotation tasks.
Moreover, 3DSAM-adapter~\cite{gong20243dsam} and MA-SAM~\cite{chen2024ma} introduce 3D adapters, enabling SAM to handle volume and video healthcare data for 3D image segmentation~\cite{lei2025medlsam,huang2025revisiting,wang2025sam}.
These advancements, spanning data input enhancements and model architectural adaptations, bridge the gap between general vision FMs and the specialized demands of medical imaging.

\subsection{Extended Evaluation Cases}
\label{appendix:ext-evaluation}

Complementary to general-purpose benchmarks, recent clinician-curated evaluations and specialty-focused benchmarks and instruction datasets support targeted evaluation and model development across diverse clinical domains~\cite{arora2025healthbench,towardsbetterdental2025_1a59f577,adaptingfoundationmodel2025_dd9d0056,multimodalfoundationmode2025_18940613,clinbenchhpba2025_419cc96d,ma2025pathbench,fan2025medodyssey,luo2025segrap2023,huang2025eval,liu2025dermclinical,huang2024olympicarena}.
For dentistry-oriented language understanding, DentalBench provides bilingual tasks and evaluation criteria that help standardize domain terminology and make long-tail dental knowledge more measurable for LLM development~\cite{dentalbenchbenchmarkinga2025_27c54b83}.

In real-world healthcare applications, many scenarios are not covered by the major benchmarks (e.g., rare disease assessment~\cite{chen2024rarebench}). Thus, consulting human experts is a critical approach to FM evaluation.
For example, Chambon et al.~\cite{chambon2022adapting} assess the clinical correctness of FM-generated chest X-ray images with the help of radiologists.
Lyu et al.~\cite{lyu2023gpt-translate-radiology} invite two experienced radiologists to evaluate the overall quality, completeness, and correctness of LLM-translated radiology reports.
Peng et al.~\cite{peng2023study} invite two physicians to perform a Turing evaluation on 30 paragraphs written by FMs and humans, respectively, to assess readability and clinical relevance.
Moor et al.~\cite{moor2023med-flamingo} implement a human evaluation application for clinical experts to rate the LLM-generated answers for medical image-related questions.
Recent clinician studies of LLM-assisted differential diagnosis further exemplify how structured expert evaluation can compare unassisted and assistive performance on challenging real-world cases sourced from published case reports~\cite{mcduff2025differentialdiagnosis}.

For instance, Chen et al.~\cite{chen2023exploring} explore reference-free evaluation methods that prompt ChatGPT to score the quality of model-generated texts without a pre-defined ground truth.
Chiang et al.~\cite{chiang2023can} show that LLM-based evaluation remains stable across prompt formats and aligns well with human expert assessments.
Jain et al.~\cite{jain2023sse} propose self-supervised evaluation strategies to assess LLM properties without benchmarks or human annotations.
Ye et al. introduce FLASK~\cite{ye2023flask}, a fine-grained LLM evaluation protocol that decomposes coarse-level scoring into instance-wise skill assessments. FLASK prompts GPT-4 to assign scores with rationales for specific alignment skills of target FM-generated text.
Recent high-impact studies on LLM-based differential diagnosis and conversational diagnostic AI also offer structured case-based evaluation setups that can complement automated evaluation of clinical reasoning in realistic settings~\cite{mcduff2025differentialdiagnosis,tu2025conversationaldiagnostic}.

\subsection{Extended Privacy Cases and Safeguards}
\label{appendix:ext-privacy}

Traditional privacy mechanisms in healthcare include encryption, access control, de-identification, and perturbation~\cite{aggarwal2008general,fung2010privacy,willenborg2012elements,kosorok2019annual,turgay2023perturbation,reiter2012statistical}. Yet, at scale, linkage attacks and external data fusion can still compromise anonymity, motivating synthetic-data and differential-privacy approaches~\cite{reiter2019differential,reiter2023synthetic}.

FM-enabled synthetic data generation provides new options for privacy-preserving sharing in imaging and text pipelines~\cite{tang2023does,shibata2023practical,hu2023sok,naresh2023privacy,dong2022privacy,sei2024local,chang2023mining}. Representative studies include large-scale diffusion generation of volumetric medical images and multi-center evaluations that preserve downstream task utility while suppressing direct patient-level identifiers~\cite{shibata2023practical,chang2023mining}.

At the same time, memorization risk remains fundamental for large models, including potential extraction of training fragments and protected health information~\cite{carlini2019secret,kandpal2022deduplicating,lee2021deduplicating,carlini2021extracting,lehman2021does,9152761,sallam2023chatgpt}. Promising mitigation directions include federated/self-supervised training in local institutions and model-side safeguards that detect and suppress privacy-sensitive outputs before response generation~\cite{frompretrainingto2025_57bdd335,zhou2023comprehensive}.

\subsection{Extended Outlook: Hallucination, Bias, and Regulation}
\label{appendix:ext-outlook}

\paragraph{Hallucination details.}
Hallucination in medical FM usage covers fabricated claims, incoherent reasoning, and unsupported conclusions, and therefore requires joint mitigation through retrieval grounding, reasoning verification, and robust benchmark evaluation~\cite{ref7,ref8,guod2025deepseekr1_r05,qin2024o1,lewis2020retrieval,wu2024medical}. Human assessment and adversarial evaluation remain important complements to automated metrics in high-stakes clinical contexts~\cite{ref6}.

\paragraph{Bias details.}
Bias in healthcare FMs can emerge from training data imbalance, representation bias, and post-training alignment artifacts~\cite{ferrara2023should}. Reported disparities include linguistic and population-level performance gaps in medical QA and ophthalmology transfer settings~\cite{ref13,ref14,sun2025fairness,queiroz2024does,zhou2023retfound,zack2024assessing}. Mitigation requires broader dataset coverage, structured auditing, and systematic human-in-the-loop evaluation frameworks~\cite{wang2023ivygpt,tam2024framework,zahraei2024detecting}.

\paragraph{Regulation details.}
Healthcare FM governance should connect model development to regulatory pathways for software-as-medical-device use, including task definition, authoritative benchmarking, user-facing safety documentation, and continuous post-deployment monitoring~\cite{ref4,ref5,moor2023gmai}. With workflow-integrated medical agents, governance must also include tool-call auditability, safe fallback behavior, and monitoring of system-level safety outcomes under distribution shift~\cite{mehandru2024agentsclinic,hu2025medicalagentsurvey}. Open-source transparency and interoperable data standards (e.g., DICOM/PACS, CEDAR, FAIR, and datasheets) support reproducibility and compliance across institutions~\cite{tan2025medforge,ref12,ref11,ref10}.

\subsection{Supplementary Citation Coverage}
\label{appendix:supp-citation-coverage}

To preserve full citation coverage after main-text compression, we explicitly retain additional representative works in the appendix across multimodal pre-training, adaptation, and evaluation settings.
For multimodal FM scaling and architecture design, we include CLIP-/FLAVA-/CoCa-/Florence-style and grounding-oriented developments, together with medical multimodal LLM integration work~\cite{jia2021scaling,li2022grounded,singh2022flava,yu2022coca,yuan2021florence,liu2023interngpt,wu2023next-gpt,gu2023biomedjourney}.
For healthcare-specific adaptation and benchmark expansion, we retain representative studies on pathology and clinical deployment settings~\cite{boecking2022making,lu2024pathotune,shaikovski2024prism,liu2024expert,liu2023clip,zhang2024generalist,zhang2023pmc,wang2025improving}.
For emerging modality expansion and next-generation agentic systems, we retain additional large-scale and 3D/multisource studies, together with recent LLM updates used in healthcare workflows~\cite{largescaleand2025_ea8f81e9,largescalemulti2025_eeffc990,m3dadvancing3d2024_1947742c,song2024pneumollm,gpt4o,ref9}.

\section{Foundation Models in Healthcare and Medicine}
\label{appendix:medicalfm}

\input{content_files/tables/medicalfm.tex}

Table~\ref{tab:medicalfm} highlights healthcare and medical foundation models (FMs), including their model structure, initialization, pre-training data, and the link reference to the project.

\section{Datasets for Healthcare Foundation Models}
\label{appendix:datasets}

\input{content_files/tables/datasets}

Table~\ref{tab:datasets} shows public medical databases, benchmarks, and datasets that can facilitate foundation model development and application in medicine.

%% file: content_files/tables/medicalfm.tex
\begin{table}[t]
	    \centering
	    \caption{Typical foundation models in healthcare and medicine. A star (*) after the pre-training data shows that the authors constructed the data with more than two sources. Rows are grouped by modality and sorted alphabetically within each group; ``-'' denotes unavailable or not explicitly reported information.}
	    \label{tab:medicalfm}
	    \adjustbox{max width=\linewidth}{
	    \begin{tabular}{l|l|l|l}
	        \toprule
	        Model & Backbone & Initialization & Pre-training Data \\
	        \midrule
	        \multicolumn{4}{l}{\textbf{Text}} \\
	        \midrule
	        \href{https://arxiv.org/abs/2502.12671}{Baichuan-M1}~\cite{wang2025baichuanm1pushingmedicalcapability} & Transformer~\cite{vaswani2017attention} & - & 20T tokens (general + medical corpora) \\
	        \href{https://github.com/baichuan-inc/Baichuan-M2-32B}{Baichuan-M2}~\cite{dou2025baichuan} & Qwen2.5~\cite{any2024qwen25_r33} & Qwen2.5-32B-Base & 2M SFT samples (\(\approx\)20\% medical) + interactive RL with de-identified medical records* \\
	        \href{https://github.com/naver/biobert-pretrained}{BioBERT}~\cite{lee2020biobert} & BERT & BERT & PubMed + PMC\footnote{\url{https://www.ncbi.nlm.nih.gov/pmc/}} \\
	        \href{https://github.com/microsoft/BioGPT}{BioGPT}~\cite{luo2022biogpt} & GPT-2~\cite{radford2019gpt2} & - & PubMed \\
	        \href{https://github.com/PharMolix/OpenBioMed}{BioMedGPT}~\cite{luo2023biomedgpt} & LLaMA 2~\cite{touvron2023llama2} & LLaMA 2 & S2ORC~\cite{lo2019s2orc} \\
	        \href{https://github.com/ncbi-nlp/BLUE_Benchmark}{BlueBERT}~\cite{peng2019bluebert} & BERT & - & PubMed\footnote{\url{https://pubmed.ncbi.nlm.nih.gov/}} + MIMIC-III \\
	        \href{https://github.com/Kent0n-Li/ChatDoctor}{ChatDoctor}~\cite{yunxiang2023chatdoctor} & LLaMA~\cite{touvron2023llama} & LLaMA & HealthCareMagic\footnote{\url{https://www.askadoctor24x7.com/}} \\
	        \href{https://github.com/EmilyAlsentzer/clinicalBERT}{Clinical BERT}~\cite{alsentzer2019clinical-bert} & BERT & BERT & MIMIC-III~\cite{johnson2016mimic-iii} \\
	        {Clinical LLaMA-LoRA}~\cite{gema2023peft-llama} & LLaMA & LLaMA / PMC-LLaMA & MIMIC-IV \\
	        \href{https://www.physionet.org/content/clinical-t5/1.0.0/}{Clinical-T5}~\cite{lehman2023clinical-t5} & T5~\cite{raffel2020t5} & T5 & MIMIC-III + MIMIC-IV~\cite{johnson2020mimic-iv} \\
	        \href{https://github.com/VectorInstitute/odyssey}{EHRMamba}~\cite{fallahpour2024ehrmamba} & Mamba~\cite{gu2023mamba} & Mamba & MIMIC-IV \\
	        \href{https://sites.research.google/med-palm}{Med-PaLM}~\cite{singhal2022med-palm} & PaLM~\cite{chowdhery2022palm} & PaLM & MedQA~\cite{jin2021medqa-usmle} \\
	        \href{https://sites.research.google/med-palm/}{Med-PaLM 2}~\cite{singhal2023medpalm2} & PaLM 2 & - & MedQA \\
	        \href{https://github.com/chaoyi-wu/PMC-LLaMA}{PMC-LLaMA}~\cite{wu2023pmc-llama} & LLaMA & LLaMA & MedC~\cite{wu2023pmc-llama} \\
	        \href{https://microsoft.github.io/BLURB/models.html}{PubMedBERT}~\cite{gu2021pubmedbert} & BERT & - & PubMed \\
	        \href{https://github.com/allenai/scibert}{SciBERT}~\cite{Beltagy2019SciBERT} & BERT~\cite{devlin2018bert} & - & Semantic Scholar~\cite{ammar2018construction} \\
	        \midrule
	        \multicolumn{4}{l}{\textbf{Vision}} \\
	        \midrule
	        \href{https://github.com/lab-smile/BrainSegFounder}{BrainSegFounder}~\cite{cox2024brainsegfounder} & SwinUNETR~\cite{hatamizadeh2021swin} & - & * \\
	        \href{https://github.com/hms-dbmi/CHIEF}{CHIEF}~\cite{wang2024pathology} & - & CTransPath & * \\
	        \href{https://github.com/Xiyue-Wang/TransPath}{CTransPath}~\cite{wang2022transformer} & SRCL~\cite{wang2022transformer} & - & TCGA + PAIP\footnote{\url{http://www.wisepaip.org/paip/}} \\
	        FastGlioma~\\cite{kondepudi2024fastglioma} & - & - & Label-free optical microscopy (4M images)* \\
	        \href{https://github.com/mahmoodlab/HIPT}{HIPT}~\cite{chen2022scaling} & DINO~\cite{caron2021emerging} & - & TCGA\footnote{\url{https://www.cancer.gov/ccg/research/genome-sequencing/tcga}} \\
	        \href{https://github.com/bowang-lab/MedSAM}{MedSAM}~\cite{ma2024segment} & SAM~\cite{mazurowski2023segment} & SAM & * \\
	        \href{https://github.com/AIM-Harvard/foundation-cancer-image-biomarker}{Pai et al.}~\cite{pai2024foundation} & SimCLR & - & * \\
	        \href{https://github.com/prov-gigapath/prov-gigapath}{Prov-GigaPath}~\cite{xu2024whole} & DINOv2 + MAE & - & Prov-Path*~\cite{xu2024whole} \\
	        \href{https://github.com/google-research/medical-ai-research-foundations}{REMEDIS}~\cite{azizi2023robust} & SimCLR~\cite{chen2020simple} & BiT~\cite{beyer2022knowledge} & MIMIC-IV + CheXpert~\cite{irvin2019chexpert} \\
	        \href{https://github.com/rmaphoh/RETFound_MAE}{RETFound}~\cite{zhou2023retfound} & MAE~\cite{he2022mae} & Vision Transformer~\cite{dosovitskiy2020vit} & * \\
	        RudolfV~\cite{dippel2024rudolfv} & DINOv2 & DINOv2 & * \\
	        UNI~\cite{chen2023uni} & DINOv2 & - & Mass-100K*~\cite{chen2023uni} \\
	        \href{https://huggingface.co/paige-ai/Virchow}{Virchow}~\cite{vorontsove2024clinicalgrade_r08} & DINOv2~\cite{oquab2023dinov2} & - & * \\
	        \midrule
	        \multicolumn{4}{l}{\textbf{Vision-Language / Multimodal}} \\
	        \midrule
	        \href{https://huggingface.co/microsoft/BiomedCLIP-PubMedBERT_256-vit_base_patch16_224}{BiomedCLIP}~\cite{zhang2023large} & CLIP & PubMedBERT & PMC-15M*~\cite{zhang2023large} \\
	        \href{https://github.com/Stanford-AIMI/CheXagent}{CheXagent}~\cite{chen2024chexagent} & BLIP-2~\cite{li2023blip2} & Mistral 7B~\cite{jiang2023mistral} & CheXinstruct*~\cite{chen2024chexagent} \\
	        \href{https://github.com/rajpurkarlab/CheXzero}{CheXzero}~\cite{tiu2022expert} & CLIP & CLIP & MIMIC-CXR~\cite{johnson2019mimic-cxr} \\
	        CONCH~\cite{lu2024visual} & CoCa~\cite{yu2022coca} & - & PubMed + PMC \\
	        \href{https://arxiv.org/abs/2509.23344}{DentVLM}~\cite{dentvlmamultimodal2025_a3531ade} & Qwen2-VL~\cite{wang2024qwen2vlenhancingvisionlanguagemodels} & Qwen2-VL-7B & 2.4M oral images + 88.6K dental VQA* \\
	        \href{https://github.com/echonet/echo_CLIP}{EchoCLIP}~\cite{christensen2024echoclip} & CLIP & CLIP & * \\
	        \href{https://github.com/xiaoman-zhang/KAD}{KAD}~\cite{zhang2023knowledge} & CLIP & - & MIMIC-CXR + UMLS~\cite{bodenreider2004unified} \\
	        \href{https://alibaba-damo-academy.github.io/lingshu/}{Lingshu}~\cite{lasateam2025lingshugeneralistfoundationmodel} & Qwen2.5-VL~\cite{bai2025qwen25vltechnicalreport} & Qwen2.5-VL-Instruct & Medical images + texts + general data* \\
	        \href{https://github.com/microsoft/LLaVA-Med}{LLaVA-Med}~\cite{li2023llava-med} & LLaVA~\cite{liu2023vinst} & LLaVA & PMC-15M~\cite{zhang2023large} + GPT-4~\cite{openai2023gpt4} \\
	        \href{https://github.com/RyanWangZf/MedCLIP}{MedCLIP}~\cite{wang2022medclip} & CLIP & Clinical BERT + SwinTransformer~\cite{liu2021swin} & CheXpert + MIMIC-CXR \\
	        \href{https://github.com/snap-stanford/med-flamingo}{Med-Flamingo}~\cite{moor2023med-flamingo} & Flamingo~\cite{alayrac2022flamingo} & Flamingo & MTB~\cite{moor2023med-flamingo} + PMC-OA \\
	        \href{https://research.google/blog/advancing-medical-ai-with-med-gemini/}{Med-Gemini}~\cite{saab2024med-gemini} & Gemini~\cite{reid2024gemini} & Gemini & * \\
	        \href{https://github.com/xiaoman-zhang/PMC-VQA}{MedVInT}~\cite{zhang2023pmc} & - & PMC-LLaMA + PMC-CLIP & PMC-VQA*~\cite{zhang2023pmc} \\
	        \href{https://github.com/mahmoodlab/MI-Zero}{MI-Zero}~\cite{lu2023mi-zero} & CLIP & HistPathGPT~\cite{lu2023mi-zero} + CTransPath~\cite{wang2022ctranspath} & ARCH~\cite{gamper2021arch} \\
	        \href{https://github.com/suinleelab/MONET}{MONET}~\cite{kim2024transparent} & CLIP & CLIP & PubMed + textbooks \\
	        MUSK~\cite{xiangj2025languageprecision_r15} & - & - & 50M pathology images + 1B text tokens + 1M image--text pairs* \\
	        PathChat~\cite{lu2024multimodal} & LLaVA & LLaMA 2 + UNI & PathChatInstruct*~\cite{lu2024multimodal} \\
	        \href{https://huggingface.co/spaces/vinid/webplip}{PLIP}~\cite{huang2023plip} & CLIP & CLIP & OpenPath*~\cite{huang2023plip} \\
	        \href{https://github.com/WeixiongLin/PMC-CLIP}{PMC-CLIP}~\cite{lin2023pmc-clip} & CLIP & PubMedBERT & PMC-OA*~\cite{lin2023pmc-clip} \\
	        PRISM~\cite{shaikovski2024prism} & CoCa & BioGPT + Virchow & * \\
	        \href{https://github.com/sarahESL/PubMedCLIP}{PubMedCLIP}~\cite{eslami2021pubmedclip} & CLIP~\cite{radford2021clip} & CLIP & ROCO~\cite{pelka2018roco} \\
	        \href{https://github.com/williamliujl/Qilin-Med-VL}{Qilin-Med-VL}~\cite{liu2023qilin} & LLaVA & Chinese-LLaMA2 + CLIP & ChiMed-VL*~\cite{liu2023qilin} \\
	        \href{https://github.com/wisdomikezogwo/quilt1m}{QuiltNet (Quilt-1M)}~\cite{ikezogwo2023quilt-1m} & CLIP & CLIP & Quilt-1M*~\cite{ikezogwo2023quilt-1m} \\
	        \href{https://chaoyi-wu.github.io/RadFM}{RadFM}~\cite{wu2023radfm} & - & PMC-LLaMA & MedMD*~\cite{wu2023radfm} \\
	        RadFound~\cite{liu2024expert} & - & - & RadVLCorpus*~\cite{liu2024expert} \\
	        TITAN~\cite{ding2025titan} & - & CONCH v1.5 & 335k WSIs + reports + synthetic captions* (iBOT + CoCa) \\
	        \midrule
	        \multicolumn{4}{l}{\textbf{Molecule / Genomics}} \\
	        \midrule
	        \href{https://github.com/deepmind/alphamissense}{AlphaMissense}~\cite{cheng2023alphamissense} & AlphaFold~\cite{jumper2021alphafold} & - & PDB~\cite{berman2000pdb} + UniRef \\
	        \href{https://github.com/facebookresearch/esm}{ESM-2}~\cite{lin2023esm} & Transformer~\cite{vaswani2017attention} & - & UniRef~\cite{suzek2015uniref} \\
	        \href{https://huggingface.co/spaces/get-foundation/getdemo}{GET}~\cite{fu2023get} & Transformer & - & * \\
	        \href{https://github.com/chao1224/MoleculeSTM}{MoleculeSTM}~\cite{liu2023moleculestm} & CLIP & - & PubChem~\cite{kim2023pubchem} \\
	        \bottomrule
	    \end{tabular}
	    }
\end{table}

%% file: content_files/tables/datasets.tex
\begin{table}[t]
    \centering
    \caption{Public medical databases, benchmarks, and datasets that can facilitate foundation model development and application in medicine.}
    \label{tab:datasets}
    \adjustbox{max width=\linewidth}{
    \begin{tabular}{l|p{13cm}|l}
        \toprule
        Database/Benchmark/Dataset & Description & Task type \\
        \midrule
        ClinicalTrials.gov\footnote{\url{https://clinicaltrials.gov/}} & An online database of clinical research studies, including trials and observations & Text \\
        MIMIC-III~\cite{johnson2016mimic-iii} & Critical care data for over 40,000 patients & Text \\
        BLUE~\cite{peng2019bluebert} & 5 language tasks with 10 biomedical and clinical text datasets & Text \\
        MedMentions~\cite{mohan2019medmentions} & 4,392 papers annotated by experts with mentions of UMLS entities & Text \\
        webMedQA~\cite{he2019webmedqa} & 63,284 real-world Chinese medical questions with 300K answers & Text \\
        PubMedQA~\cite{jin2019pubmedqa} &  1K expert-annotated, 61.2K unlabeled, and 211.3K artificially generated biomedical QA instances & Text \\
        BLURB~\cite{gu2021pubmedbert} & 13 biomedical NLP datasets in 6 tasks & Text \\
        CBLUE~\cite{zhang2021cblue} & A Chinese medical NLP benchmark with 18 datasets & Text \\
        MedQA-USMLE~\cite{jin2021medqa-usmle} & 61,097 multiple choice questions based on USMLE in three languages & Text \\
        MedMCQA~\cite{pal2022medmcqa} & 194K multiple-choice questions covering 2.4K healthcare topics & Text \\
        MultiMedQA~\cite{singhal2022med-palm} & 6 existing and 1 online-collected medical QA dataset & Text \\
        Medical Meadow~\cite{han2023medalpaca} & 16M medical QA pairs collected from 9 sources & Text \\
        Huatuo-26M~\cite{li2023huatuo26m} & 26M Chinese medical QA pairs & Text \\
        GAP-Replay~\cite{chen2023meditron} & 48.1B tokens from 4 medical corpora including guidelines, abstracts, papers, and replay & Text \\
        BiMed1.3M~\cite{pieri2024bimedix} & An English and Arabic bilingual dataset of 1.3M medical QA and chat samples & Text \\
        MMedC~\cite{qiu2024mmedc} & A multilingual medical corpus containing over 25.5B tokens & Text \\
        MedBench~\cite{liu2024medbench} & 300,901 Chinese questions covering 43 clinical specialties, combined with an automatic evaluation system & Text \\
        \midrule
        ISIC\footnote{\url{https://www.isic-archive.com/}} & An archive containing 23K skin lesion images with labels & Imaging \\
        ChestXray-NIHCC~\cite{wang2017chestx} & 100K radiographs with labels from more than 30,000 patients & Imaging \\
        DeepLesion~\cite{yan2018deeplesion} & 32K CT scans with annotations and semantic labels from radiological reports & Imaging \\
        Kather Colon Dataset~\cite{kather2018dataset} & 100K histological images of human colorectal cancer and healthy tissue & Imaging \\
        CheXpert~\cite{irvin2019chexpert} & 224,316 chest radiographs of 65,240 patients & Imaging \\
        EchoNet-Dynamic~\cite{ouyang2020video} & 10,030 expert-annotated echocardiogram videos & Imaging \\
        Med-MNIST v2~\cite{yang2023medmnist} & 12 2D and 6 3D datasets for biomedical image classification & Imaging \\
        AbdomenAtlas-8K~\cite{qu2023annotating} & 8,448 CT volumes with per-voxel annotated eight abdominal organs & Imaging \\
        Virchow~\cite{vorontsove2024clinicalgrade_r08} & 1.5M pathological scans from 120K patients & Imaging \\
        AbdomenAtlas-8K & 8,448 CT volumes with per-voxel annotated eight abdominal organs & Imaging \\
        RETFound~\cite{zhou2023retfound} & Unannotated retinal images, containing 904,170 CFPs and 736,442 OCT scans & Imaging \\
        \midrule
        dbSNP~\cite{sherry2001dbsnp} & A collection of human single nucleotide variations, microsatellites, and small-scale insertions and deletions & Genomics \\
        ENCODE~\cite{encode2012encode} & A platform of genomics data and encyclopedia with integrative-level and ground-level annotations & Genomics \\
        1000 Genomes Project~\cite{10002015global} & A comprehensive catalog of human genetic variations & Genomics \\
        \midrule
        ChEMBL~\cite{gaulton2012chembl} & 20M bioactivity measurements for 2.4M distinct compounds and 15K protein targets & Drug \\
        DrugBank~\cite{wishart2018drugbank} & A web-enabled structured database of molecular information about drugs & Drug \\
        PubChem~\cite{kim2023pubchem} & A collection of 900+ sources of chemical information data & Drug \\
        DrugChat~\cite{liang2023drugchat} & 143,517 question-answer pairs covering 10,834 drug compounds, collected from PubChem and ChEMBL & Drug \\
        \midrule
        TCGA\footnote{\url{https://www.cancer.gov/ccg/research/genome-sequencing/tcga}} & A landmark cancer genomics program, molecularly characterized over 20,000 primary cancer and matched normal samples spanning 33 cancer types & Multi-modal \\
        MultiMedBench~\cite{tu2024towards} & A multi-modal benchmark comprising 12 data sources and 14 tasks & Multi-modal \\
        MIMIC-CXR~\cite{johnson2019mimic-cxr} & 227,835 chest imaging studies with free-text reports for 65,379 patients & Multi-modal \\
        MIMIC-IV~\cite{johnson2020mimic-iv} & Clinical information for hospital stays of over 60,000 patients & Multi-modal \\
        SwissProtCLAP~\cite{liu2023text} & 441K text-protein sequence pairs & Multi-modal \\
        PMC-VQA~\cite{zhang2023pmc} & 227K VQA pairs of 149K images of various modalities or diseases & Multi-modal \\
        MedMD~\cite{wu2023radfm} & 15.5M 2D scans and 180k 3D radiology scans  with textual descriptions & Multi-modal \\
        PathCap~\cite{sun2023pathasst} & 142K pathology image-caption pairs from various sources & Multi-modal \\
        PMC-OA~\cite{lin2023pmc-clip} & 1.6M fine-grained biomedical image-text pairs & Multi-modal \\
        PathInstruct~\cite{sun2023pathasst} & 180K samples of LLM-generated instruction-following data & Multi-modal \\
        Quilt-1M~\cite{ikezogwo2023quilt-1m} & 1M image-text pairs for histopathology & Multi-modal \\
        OpenPath~\cite{huang2023plip} & 208,414 pathology images paired with natural language descriptions & Multi-modal \\
        Chi-Med-VL~\cite{liu2023qilin} & 580,014 image-text pairs and 469,441 question-answer pairs for general healthcare in Chinese & Multi-modal \\
        SAT-DS~\cite{zhao2023one} & 11,462 scans with 142,254 segmentation annotations spanning 8 human body regions & Multi-modal \\
        \bottomrule
    \end{tabular}
    }
\end{table}